\documentclass[runningheads]{llncs}
\usepackage{graphicx}

\usepackage{tikz}
\usepackage{comment}
\usepackage{amsmath,amssymb} 
\usepackage{color}

\usepackage[accsupp]{axessibility}  

\usepackage{graphicx}
\usepackage{bbding}
\usepackage{float} 
\usepackage{pifont}
\usepackage{booktabs}
\usepackage{multirow}
\usepackage{multicol}
\usepackage{algpseudocode}

\usepackage{subfigure}

\usepackage[capitalize]{cleveref}
\crefname{section}{Sec.}{Secs.}
\Crefname{section}{Section}{Sections}
\Crefname{table}{Table}{Tables}
\crefname{table}{Tab.}{Tabs.}

\DeclareMathOperator*{\argmax}{arg\,max}

\def\eg{\emph{e.g. }}

\def\ie{\emph{i.e. }}

\def\vs{\emph{vs. }}

\makeatletter
\def\thanks#1{\protected@xdef\@thanks{\@thanks
        \protect\footnotetext{#1}}}
\makeatother

\newcommand{\PAR}[1]{\smallskip \noindent \textbf{#1}}

\newcommand{\figref}[1]{Figure~\ref{#1}}
\newcommand{\tabref}[1]{Table~\ref{#1}}
\renewcommand{\eqref}[1]{Eq.~(\ref{#1})}

\begin{document}
\pagestyle{headings}
\mainmatter
\def\ECCVSubNumber{1588}  

\title{TransFGU: A Top-down Approach to Fine-Grained Unsupervised Semantic Segmentation} 

\titlerunning{TransFGU}
%
\author{Zhaoyuan Yin\inst{1,2}$^{\ast}$\thanks{
* Work done during an internship at Alibaba Group.\\
\dag~Corresponding author, project lead.\\
\ddag~Corresponding author.
}\index{Zhaoyuan Yin} 
\and Pichao Wang\inst{2\dag}
\and Fan Wang\inst{2}
\and Xianzhe Xu\inst{2}
\and \\ Hanling Zhang\inst{3\ddag}
\and Hao Li\inst{2} 
\and Rong Jin\inst{2}}
\authorrunning{Zhaoyuan Yin et al.}
%

\institute{College of Computer Science and Electronic Engineering, Hunan University, China \email{zyyin@hnu.edu.cn} \and Alibaba Group, China\\ \email{\{pichao.wang, fan.w, xianzhe.xxz, lihao.lh, jinrong.jr\}@alibaba-inc.com} \and School of Design, Hunan University, China\\ \email{jt\_hlzhang@hnu.edu.cn}\\}


\maketitle

\begin{abstract}

Unsupervised semantic segmentation aims to obtain high-level semantic representation on low-level visual features without manual annotations. Most existing methods are bottom-up approaches that try to group pixels into regions based on their visual cues or certain predefined rules. As a result, it is difficult for these bottom-up approaches to generate fine-grained semantic segmentation when coming to complicated scenes with multiple objects and some objects sharing similar visual appearance. In contrast, we propose the first top-down unsupervised semantic segmentation framework for fine-grained segmentation in extremely complicated scenarios. Specifically, we first obtain rich high-level structured semantic concept information from large-scale vision data in a self-supervised learning manner, and use such information as a prior to discover potential semantic categories presented in target datasets. Secondly, the discovered high-level semantic categories are mapped to low-level pixel features by calculating the class activate map (CAM) with respect to certain discovered semantic representation. Lastly, the obtained CAMs serve as pseudo labels to train the segmentation module and produce the final semantic segmentation. Experimental results on multiple semantic segmentation benchmarks show that our top-down unsupervised segmentation is robust to both object-centric and scene-centric datasets under different semantic granularity levels, and outperforms all the current state-of-the-art bottom-up methods. Our code is available at \url{https://github.com/damo-cv/TransFGU}.

\end{abstract}

\section{Introduction}
\label{sec:intro}

\begin{figure}[t]
    \centering
    \begin{minipage}[c]{0.67\textwidth}
        \includegraphics[width=\linewidth]{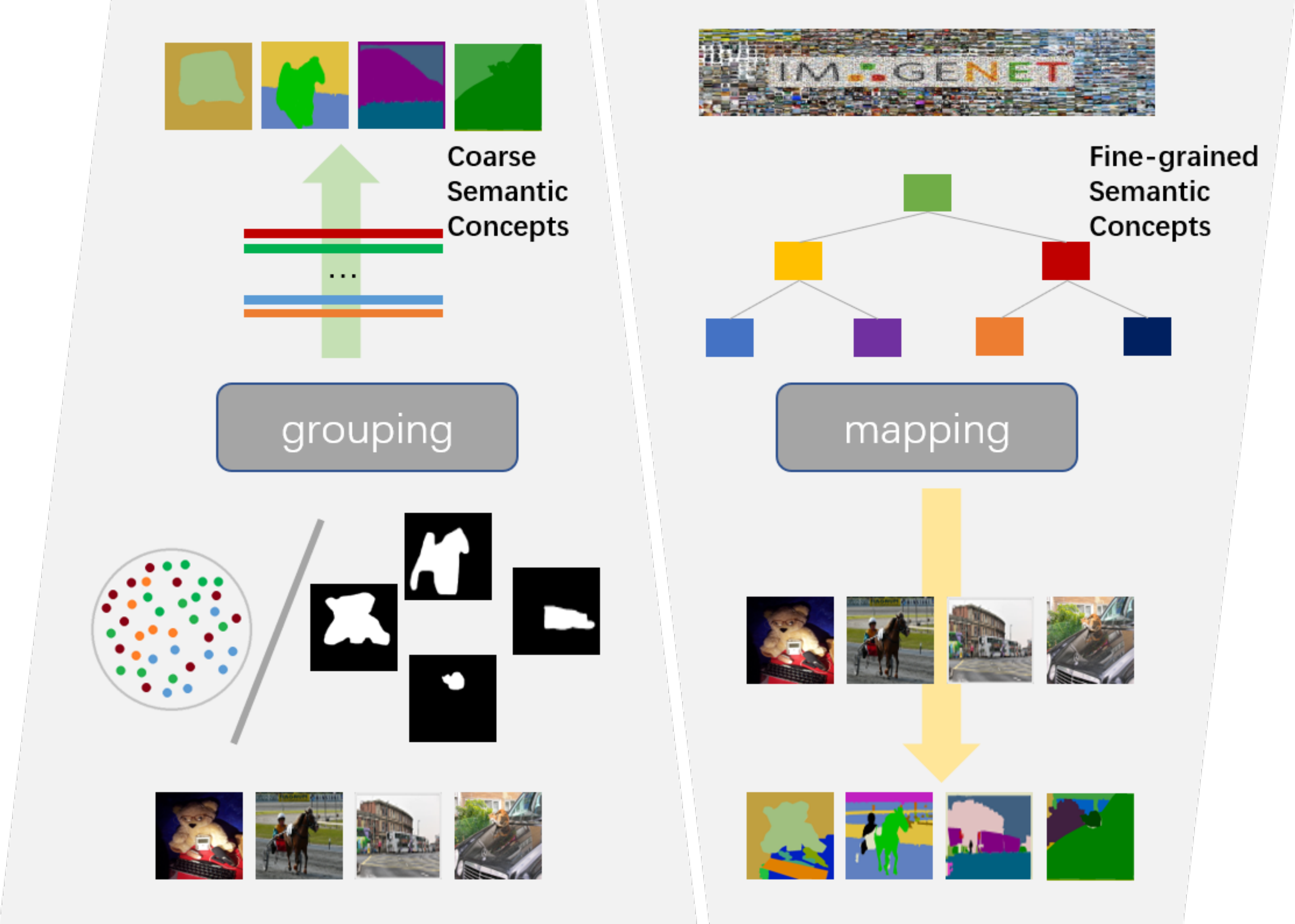}
    \end{minipage}\hfill
    \begin{minipage}[c]{0.3\textwidth}
        \caption{Bottom-up (left) \vs Top-down (right) frameworks. Currently bottom-up manners group feature under the guidance of certain prior knowledge to form the semantic concepts (usually coarse), while our top-down manner maps the fine-grained structural semantic concepts obtained from ImageNet to pixel-level features and generates fine-grained segmentation.\label{fig:teaser}}
    \end{minipage}
\end{figure}

Given pictures of this world, can a semantic concept be deduced by certain prior rules, or can it be induced from the massive amount of observations? The answer to this question leads to different ways to obtain the semantic concept, and therefore brings different paradigms to the task of unsupervised semantic segmentation, which aims to obtain the pixel-wise classification as the semantic concept without any manual-annotated labels.

One way to tackle unsupervised image segmentation is to group low-level pixels into some semantic groups under the guidance of certain prior knowledge, \ie the bottom-up manner~\cite{PiCIE,IIC,InfoSeg,kanezaki2018unsupervised,kim2020unsupervised,InMARS,MaskContrast}, as shown in Figure~\ref{fig:teaser}. Those methods often assume pixels in the same semantic object share a similar representation in the high-level semantic space. However, there is a large gap between low-level pixel and high-level semantic embeddings. Two semantically different objects may be mostly similar in low-level feature space, while the key to distinguishing them lies in some small areas reflecting the uniqueness of a semantic category. \eg, the difference between a horse and a donkey may be found only on ears and legs. Accurate perception of these slight differences is the key to generating fine-grained segmentation, which is very hard to obtain by pixel-level deduction. In contrast, an object that has a large intra-class variance in appearance, \eg the different parts of a person (head, arms, body...), may lead to dissimilar pixel-level features in different parts, which hinder the bottom-up methods from grouping these dissimilar features as an integrated object in high-level semantic space, due to the lack of high-level conceptual understanding of the whole object.

To alleviate these problems, this work proposes a top-down approach to unsupervised semantic segmentation, as shown in \figref{fig:pipeline}. Instead of deducing high-level semantic concepts from low-level visual features, we start from the high-level fine-grained semantic knowledge induced from ImageNet. We benefit from the self-supervised learning method DINO~\cite{DINO} to gain the initial segmentation property of self-attention maps. The semantic representation obtained from DINO is more robust to the object appearance variations. Then we leverage the obtained prior to discover all the potential semantic categories presented in the target dataset, and group them into the desired number of semantic clusters according to their semantic similarity. It makes our method flexible to semantic concepts at different granularity levels. We then project the high-level semantic information to low-level pixel feature space by Grad-CAM~\cite{selvaraju2017gradcam,zou2020pseudoseg,shi2021zoomcam}, to generate fine-grained active maps for various semantic classes. The obtained active maps serve as pseudo labels for segmentation model training. A bootstrapping mechanism is employed to iteratively refine the quality of pseudo labels and gradually improve the final segmentation performance. The proposed method, named TransFGU, bridges the gap between high-level semantic representation induced by SSL and low-level pixel space, resulting in fine-grained segmentation in both object-centric and scene-centric images without any human annotations.

Through experiments on four public benchmarks, \ie MS-COCO~\cite{lin2014mscoco}, PascalVOC~\cite{everingham2010pascalvoc}, Cityscapes~\cite{cordts2016cityscapes}, and LIP~\cite{gong2017lip}, we show that our method can handle the cases of both  foreground-only segmentation  and foreground/background segmentation. Moreover, our method can control the semantic granularity level of segmentation by adjusting hyper-parameters, which overcome the limitations of previous methods that they can only produce coarse-level segmentation results. Our method achieves state-of-the-art results on all the benchmarks, surpassing all the current bottom-up methods.

In summary, our contributions are as follows: (i) We propose the first top-down framework for unsupervised semantic segmentation, which directly exploits the semantic information induced from ImageNet to tackle the fine-grained segmentation in an annotation-free way. (ii) We design an effective mechanism that successfully transfers the high-level semantic features obtained from SSL into low-level pixel-wise features to produce high-quality fine-grained segmentation results. (iii) Experiments show that our proposed method achieves state-of-the-art on a variety of semantic segmentation benchmarks including  MS-COCO, Pascal-VOC, Cityscapes, and LIP.

\section{Related Work}
\label{sec:related_work}

\subsection{Self-supervised Representation Learning}
Self-supervised representation learning has rapidly gained attention for its promising ability to represent the feature of semantic concepts without additional labels. It is also beneficial to many downstream tasks including detection, segmentation, and tracking. The common paradigm of self-supervised representation learning is to minimize the feature distance between two views of the same object-centric image\cite{SwAV,SimSiam,SimCLR,BYOL,MoCoV1,MoCoV2,MoCoV3,EsViT,DINO}. It allows learning semantic object concepts purely based on images from ImageNet without any annotations. The learned feature can perform well with KNN, which indicates the structural-semantic concepts have been exploited. BYOL~\cite{BYOL} shows that representation learning without a single label can be on par or even better than its supervised counterpart. DINO~\cite{DINO}, a recent work based on the Transformer, shows that the self-supervised learning method can learn a good representation of structured semantic category information, which has a nice property of focusing on the area of foreground objects. While most of them aim to learn the overall representation from the object-centric image, some others\cite{DenceCL,PixPro,DSC,VADeR,SetSim,VFS} tend to generate pixel-level dense features, which better benefit the tasks that require dense matching like segmentation or detection. However, all of these methods tend to learn the pixel-level correspondence on the category of objects that appear in different views of an image rather than learning the pixel-level semantic concept, so the learned representation embedding feature cannot convert to segmentation mask directly without additional manual labels.

\subsection{Unsupervised Segmentation}

Most of the current works proposed to tackle unsupervised segmentation start from the observation of pixel-level features and grouping the pixels into different semantic groups with various priors rules that are independent of the training data. \cite{kanezaki2018unsupervised,kim2020unsupervised} group the similar pixel extracted from a randomly initialized CNN in both embedding and spatial space while keeping the diversity of embedding features. \cite{IIC,InfoSeg,InMARS,ACARL} maximize the mutual information between the pixel-level feature of two views from the same input image to distill the information shared across the image. PiCIE~\cite{PiCIE} disentangles features between different semantic objects by leveraging two simple rules, \ie invariance to geometric and equivariance to photometric while transforming two different views of an image and its feature by two asymmetric augmentation processes. These rules often fail to tackle the case with complex scenes, where objects of different categories might share similar appearances and objects of the same category might have a large intra-class appearance variation. Unlike the bottom-up manner, our method is based on the top-down pipeline that obtains fine-grained semantic concepts prior information before performing the segmentation, which is crucial to tackling these two problems.

Besides the bottom-up methods, several works use intermediate features as additional cues to guide the semantic feature grouping,  including saliency map~\cite{MaskContrast,Labels4Free,ReDO,bielski2019emergence}, super-pixel~\cite{kanezaki2018unsupervised,InMARS}, and shape prior\cite{SegSort,ShapePrior}. MaskContrast~\cite{MaskContrast} leverages two additional unsupervised saliency models to generate pseudo labels of the foreground object in each image and train the final segmentation model with a two-step bootstrapping mechanism, and then cluster all the gathered foreground masks to form the semantic concept by contrast learning. However, these methods assume that objects in an image are salient enough, which is not always true in some complicated scene-based cases and could lead to low-quality segmentation.

\begin{figure*}[t]
    \centering
    \includegraphics[width=\linewidth]{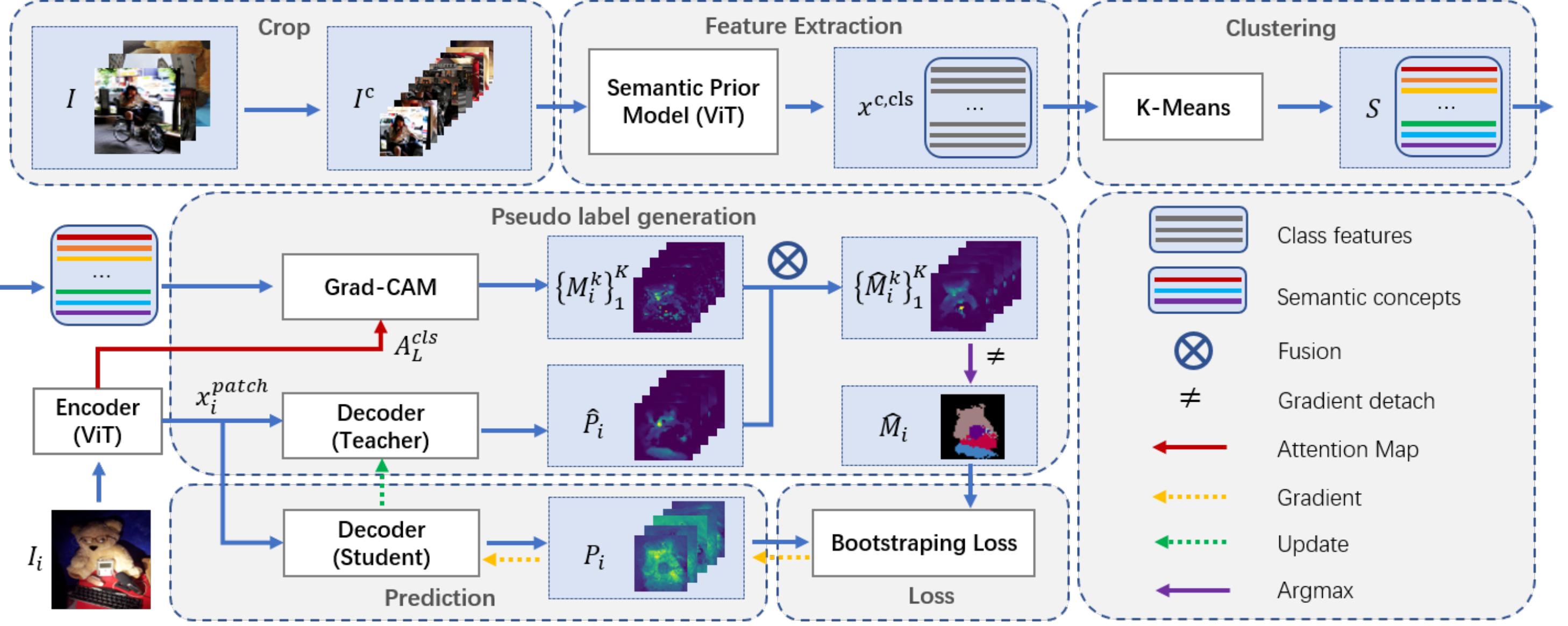}
    \caption{Our proposed top-down pipeline for unsupervised semantic segmentation. The training samples are first cropped and resized into square patches using sliding windows, and class features for each cropped patch are extracted by ViT pretrained on DINO. The obtained class features are clustered into semantic concept representations, which are mapped to the pixel level to generate the pseudo labels. The quality of pseudo labels is further bootstrapped by the teacher network, based on which a student network is trained to refine the semantic knowledge. The teacher network is updated by the learned student network to produce better pseudo labels for the next round of training. \label{fig:pipeline}}
\end{figure*}

\section{Methods}

Given a set of images $I = [I_1, ..., I_N] \in [0, 1] ^{N \times 3 \times H \times W}$ with total $N$ samples, 
our goal is to generate a group of $K$ segmentation probability maps for each image, denoted as $P = [P_1, ..., P_N] \in [0, 1]^{N \times K \times H \times W}$, where $K$ indicates the predefined number of semantic categories to be segmented in $I$, and $(H, W)$ are the height and width of the image, respectively. To achieve this goal, a semantic prior model self-supervised trained on ImageNet is introduced to extract the potential top-level semantic information from the whole set of images $I$, and the extracted top-level semantic features are clustering into $K$ groups based on their semantic similarity, resulting in $K$ semantic concept features. Then, for each $I_i \in I$, a top-down semantic information mapping pipeline is introduced to generate pixel-level semantic pseudo label according to $K$ semantic concept features based on Grad-CAM. An encoder is needed to extract pixel-level features, and a decoder is required to convert the pixel-level features into $P_i$ corresponding to the generated semantic pseudo label. In this section, we will first introduce some basics of the encoder and decoder, then describe the designed top-down pipeline, followed by a bootstrapping process that helps further refine the results.

\subsection{Revisiting ViT and Segmenter} 
Previous works on SSL have shown that ViT~\cite{dosovitskiy2020ViT}  models pretrained on ImageNet using methods like 
DINO~\cite{DINO} can provide surprisingly good properties for segmentation tasks due to its explicit semantic information learned during SSL. Thus, the pre-trained ViT is employed as $\Call{Encoder}{\cdot}$ in our pipeline, for its capability of providing high-level semantic priors through the class token, as well as extracting pixel-level features in image patch tokens. Next, we will introduce some basic notations and terminologies about ViT for easier reference.

ViT is an attention-based model with $L$ layer attention blocks. Each attention block at the $l$-th layer produces an attention map $A_l \in [0, 1]^{(1 + hw) \times (1 + hw)}$ that can be formulated as:
\begin{align}
    A_l &= \Call{Softmax}{\frac{Q_l \cdot K_l^ \mathrm{ T }}{\sqrt{d^\prime}}}
\end{align}
where $Q_l,K_l \in R^{(1 + hw) \times d^{\prime}}$ are the query and key, respectively, which are mapped and reshaped from the layer input $x_l$, $d^{\prime}$ is the dimension of embedding feature in attention block, and $(h, w)$ is the size of the feature map.

As an $\Call{Encoder}{\cdot}$, the final outputs of ViT are denoted as class token $x^{cls}$ and image patch tokens as $x^{patch}$. It is worth noting that, ViT pretrained with DINO is more irreplaceable for generating high-level semantic priors through class token, however, the model for generating pixel-level features can be substituted by other networks capable of extracting nice local features. Here we unify the two networks as the same ViT model for simplicity. 

For the decoder which generates segmentation probability maps using the output of encoder, there are also lots of options while transformer-based methods are preferable due to their robustness to noises. Segmenter~\cite{strudel2021segmenter} is selected among them for its simplicity and straightforward interface to take $x^{patch}$ as input.

Segmenter is a transformer-based segmentation model which consists of multiple cross-attention layers. It takes  $x^{patch}$  as input, and the output probability mask is: 
\begin{align}
    \Call{Mask}{x^\prime, C} = \frac{x^\prime C^\mathrm{T}}{\sqrt{d}}
    \label{eq:segmenter}
\end{align}
where $x^\prime$ and $C \in \mathcal{R}^{K \times d}$ are the patch embedding and class embedding output by the final layer in the segmenter, respectively. $\Call{Mask}{x^\prime, C} \in R^{hw \times K}$ is the downsampled probability maps for $K$ categories, and then it is upsampled to $(H, W)$ to obtain the final mask $P$. This whole process is denoted as $\Call{Decoder}{\cdot}$ in the following sections.

\subsection{Top-down pipeline}

To generate semantic mask $P$, a ViT model obtained by DINO is first used as a semantic prior model to encode all the potential semantic information in $I$. In the real-world application, an image might contain abundant and complex semantic concepts instead of a simple object, therefore, applying ViT encoder to the whole image might not be sufficient to attend to all potential semantic priors. Instead, we propose to apply a sliding window cropping operation to each image. For every image $I_i$,  $n_c$ square patches are generated, each with side length $\beta \times\min(H, W)$ where $\beta$ is a scaling factor taking a series of values. 
Patches from all images are put together to form a larger ``patch image'' set $I^c = [I^c_1, \dots, I^c_N]$ with $I^c_i$ representing the patch set for image $I_i$.
All patches are resized to  $(\frac{\min(H, W)}{2}, \frac{\min(H, W)}{2})$ and treated as a whole image before feeding into the ViT encoder.

Class features and patch features are extracted from $I^c$, denoted as $x^{c,\textit{cls}} = [x_1^{c,\textit{cls}}, \dots, x_N^{c,\textit{cls}}]$ and  $x^{c,\textit{patch}} = [x_1^{c,\textit{patch}}, \dots, x_N^{c,\textit{patch}}]$, in which:
\begin{align}
    (x^{c,\textit{cls}}_i, x^{c,\textit{patch}}_{i}) = \Call{Encoder}{I_i^c}
\end{align}
here the ViT encoder is served as a semantic prior model, $x^{c,\textit{cls}}$ can be regarded as containing all the potential semantic concepts that have appeared in $I$, $x^{c,\textit{patch}}$ is discarded because of its weak semantic prior property. Assuming there are in total $K$ pre-defined semantic categories, we can obtain  a set of semantic feature $S = [S_1, \dots, S_K] \in \mathcal{R}^{K \times d}$ by applying $K$-means on $x^{c,\textit{cls}} \in \mathcal{R}^{(N \times n_c) \times d}$ over all the $(N \times n_c)$ features based on Euclidean distance. $K$ can be set as the number of classes, to generate segmentation results at different desired granularity levels.

The extracted semantic features $S_k \in S$ can be seen as a category-wise pseudo semantic label corresponding to a certain granularity level of semantic. Next, the Grad-CAM~\cite{selvaraju2017gradcam} is adopted to visualize the corresponding category-specific response area to the target pseudo semantic label $S$. The response heat-map is treated as a coarse semantic area that locates the target semantic object in the pixel-level feature space. Furthermore, these coarse semantic areas are treated as pseudo labels $M = [M_1, \dots, M_N]$, $M_i \in [0, K]^{H \times W}$, and a decoder will be trained based on these pseudo labels.

To be more specific, inspired by \cite{chefer2021TransformerExplainability}, we first obtain the class token $x_i^{\textit{cls}}$ and patch token $x_i^\textit{patch}$ of the whole image $I_i$ by ViT encoder:
\begin{align}
    (x^{\textit{cls}}_i, x^{\textit{patch}}_{i}) = \Call{Encoder}{I_i},
\end{align}
then take attention map from $\textit{cls}$ token of the last attention block, and take the feature map while discarding the $\textit{cls}$ token itself. Denote $A_L^{\textit{cls}} \in [0, 1]^{h \times w}$ as the feature map,  gradient is generated on $A_L^{\textit{cls}}$ w.r.t $S_k$ by maximizing the cosine similarity between $x_i^{\textit{cls}}$ and each $S_k \in S$:
\begin{align}
    \min (1 - \frac{x_i^{\textit{cls}}S_k^\mathrm{T}}{\sqrt{d}}).
\end{align}
We take the gradient value of all the patch locations on $A_L^{\textit{cls}}$, and add it back to $A_L^{\textit{cls}}$ to generate $K$ response maps $[M_i^1, \dots, M_i^K] \in \mathcal{R}^{K \times h \times w}$, then the pseudo label $M_i$ can be obtained by appling $\argmax$ operator to each patch location $(h^{\prime}, w^{\prime})$ across all the $K$ response maps, where $h^{\prime} \in [0, h], w^{\prime} \in [0, w]$, and upsample to the target size of $(H, W)$:
\begin{align}
    M_i &= \Call{Upsample}{\argmax_{(h^{\prime}, w^{\prime})} (M_i^1, \dots, M_i^K)} , \\
    M_{i}^{k} &= \Call{Grad-CAM}{x_i^{\textit{cls}}, S_k | A_L^{\textit{cls}}} + A_L^{\textit{cls}}.
\end{align}
If only the foreground objects need to be segmented in an image, we calculate the background probability  $M_i^{bg}$ based on all the foreground pseudo labels as follows:
\begin{align}
    M_i^{bg} = \Call{Relu}{T^{\textit{bg}} - \max_{k \in [0, K]}{M_i^k}}
    \label{eq:prob_bg}
\end{align}
in which $T^{\textit{bg}}$ is a hyper-parameter that represents the max probability of background, and is set to 0.1 in our experiments. $M_i^{\textit{bg}}$ is then upsampled and concatenated with $M_i$. 
The decoder can be trained by standard cross-entropy loss with its output denoted as segmentation probability maps $P$:
\begin{align}
    \mathcal{L} = \Call{CE}{P, M}.
    \label{eq:loss:ce}
\end{align}

\subsection{Bootstrapping}
Training the decoder directly with the initial pseudo labels may lead to segmentation of low quality because the noise level in pseudo labels can be detrimental. For example, two categories with similar high-level semantic meaning may generate similar response maps on the same image, thus disturbing the ranking of $K$ values on the same location of the pseudo labels and further misleading the learning process. Additionally, the boundary produced by the pseudo labels is not precise enough for certain semantic objects due to the coarse response area in response maps generated by Grad-CAM. 

We tackle these problems by a bootstrapping mechanism. First, a teacher-student framework is introduced to refine the pseudo labels progressively. Second, we introduce a set of loss functions to force the model to learn discriminative abilities and prevent the model from over-fitting the noise in the pseudo labels.

\PAR{Teacher-Student Network}. In our design, both teacher and student network have exactly the same architectures as $\Call{Decoder}{\cdot}$. As shown in Figure~\ref{fig:pipeline}, given an initial pseudo mask $M$ generated by Grad-CAM, its bootstrapped version $\hat{M} = [\hat{M}_1, \dots, \hat{M}_N]$ can be obtained by aggregating $M$ and the output of the teacher network prediction $\hat{P} = [\hat{P}_1, ..., \hat{P}_N] \in [0, 1]^{N \times K \times H \times W}$:
\begin{align}
    \hat{P}_i &= \Call{Teacher}{x_{i}^{\textit{patch}}}, \\
    \hat{M}_i &= \Call{Upsample}{\argmax_{(h^{\prime}, w^{\prime})}(\hat{M}_i^1, \dots, \hat{M}_i^K)} , \\
    \hat{M}_i^k &= 0.5 \cdot (M_i^k + \hat{P}_i^k), k \in [1, \dots, K].
\end{align}
where $\hat{P}_i^k \in \mathcal{R}^{h \times w}$ is the $k$-th probability map in $\hat{P}_i$ before upsampled. With this bootstrapped pseudo labels, the student network is trained with a cross-entropy loss $\Call{CE}{P, \hat{M}}$ with $P$ as the output segmentation probability of student network, \ie $P_i = \Call{Student}{x_{i}^{\textit{patch}}}$.

In the next round, the teacher network is updated with the parameter trained from the student network, so it can produce a better prediction of $\hat{P}$, which further improves the bootstrapped pseudo labels $\hat{M}$. The student network in the next round is reset to the initial state and trained with the improved $\hat{M}$. Therefore, the quality of bootstrapped pseudo labels and the output of the student network can be improved progressively as the iteration goes on. 

\PAR{Bootstrapping Loss}. A few additional loss functions are further proposed to provide better guidance for training the decoder based on the pseudo label masks. 

First, we observe that even though categories with similar semantic meanings are difficult to differentiate thus might confuse the training process, categories with much different semantic meanings are actually easier to identify with high accuracy. Therefore, in addition to the cross-entropy loss which aligns probability masks $P$ and the ``correct'' pseudo labels $\hat{M}$, we also introduce a set of ``wrong'' pseudo labels $\hat{M}^{\prime}$ so that its cross-entropy loss with $P$ should be maximized. Thus a peer loss~\cite{liu2020PeerLoss} is defined as below:
\begin{align}
    \mathcal{L}_{peer} = \Call{CE}{P, \hat{M}} - \alpha \cdot \Call{CE}{P, \hat{M}^{\prime}}.
    \label{eq:loss:peer}
\end{align}
where $\alpha$ is a hyper-parameter, and $\hat{M}^{\prime}$ is the constructed negative labels by shuffling the $K$ pseudo labels of $\hat{M}$.

Second, it is observed that categories with similar semantic meanings usually generate similar response values at the same location of the pseudo label masks, making it hard to distinguish the correct category from the others and further slow down the training.  An uncertainty loss is introduced to diminish such uncertainty. Specifically, this is achieved by maximizing the gap between the largest and the second largest value at each location of the output probability map $P$:
\begin{align}
    \mathcal{L}_{unc} = 1 - \frac{1}{hw} \sum_{(h^{\prime}, w^{\prime})}(p^{\prime} - p^{\prime\prime}).
    \label{eq:loss:unc}
\end{align}
in which $p^{\prime}, p^{\prime\prime}$ are the largest and second largest probability value among the $K$ probabilities at location $(h^{\prime}, w^{\prime})$. 

Third, it is beneficial to keep the representations diversified in the decoder model to learn more meaningful categories information. This is done by a diversity loss which maximizes the summation of all the pairwise distances between the class embeddings of $K$ categories, \ie, minimizing their cosine similarities: 
\begin{align}
    \mathcal{L}_{div} = 1 + \frac{1}{K^2}\sum\frac{C \cdot C^T}{\sqrt{d}}.
    \label{eq:loss:div}
\end{align}
where $C$ is the class embedding in the decoder as in Eq.~\ref{eq:segmenter}.

The final loss is:
\begin{align}
    \mathcal{L} = \mathcal{L}_{peer} + \omega_1 \cdot \mathcal{L}_{div} + \omega_2 \cdot \mathcal{L}_{unc}.
    \label{eq:loss:all}
\end{align}
in which $\omega_1, \omega_2$ are hyper-parameters to balance between different losses, and set to 1 and 0.3 in our experiments, respectively.

\section{Experiments}

\subsection{Dataset and Evaluation Metrics}

We conduct experiments on four semantic segmentation benchmarks including COCO-Stuff, Pascal-VOC, Cityscapes, and LIP, with different definitions of semantic concepts.

\PAR{COCO-Stuff.} COCO-Stuff is a challenging benchmark with scene-centric images, which contains 80 things categories and 91 stuff categories. The total number of training and validation samples is 117,266 and 5,000 images, respectively. Note that the previous works~\cite{PiCIE,IIC} only conduct their experiments on the ``curated'' data (2175 images total) in which the stuff occupied at least 75\% pixels of an image. We evaluate our method on three settings: 1) COCO-S-27:all categories in COCO-Stuff merged into 27 superclasses as \cite{PiCIE}, 2) COCO-S-171: the original 171 things and stuff categories, and 3) COCO-80: only 80 things categories without background stuff.

\PAR{Cityscapes.} Cityscapes contain 5,000 images focusing on street scenes, in which 2,975 and 500 images are used for training and validation. All pixels are categorized into 34 classes. We follow \cite{PiCIE} to merge 34 classes into 27 classes after filtering out the `void' class.

\PAR{Pascal-VOC.} Pascal-VOC contains 1464 images for training and 1,449 for evaluation, with 20 classes as foreground objects to be segmented and the rest pixels as background. 

\PAR{LIP.} LIP contains person images cropped from MS-COCO~\cite{lin2014mscoco}. It has 30,462 training images and 10,000 validation images, with 19 semantic parts defined. The 19 categories are merged into 16 and 5 coarse categories to evaluate the ability of our method to handle the semantic concepts of different granularities. Please refer to supplementary materials for more details.

\PAR{Evaluation Metric.} Following the standard evaluation protocols for unsupervised segmentation used in \cite{IIC,PiCIE,MaskContrast}, we use Hungarian matching algorithm to align indices between our prediction and the ground-truth label over all the images in the validation set. Two metrics are reported for comparison, \ie Intersection over Union (mIoU) and Pixel Accuracy over all the semantic categories.

\subsection{Implementation Details} 

\PAR{Cropping and evaluation protocol}. The scaling factor $\beta$ in sliding window cropping is set to $0.5, 0.4, 0.3, 0.2$, and the step size of the sliding window is set to $\frac{1}{2} \times \beta \times \min(H, W)$. 

Foreground prior is introduced in the cropping operation, which is defined as the binarized attention map $\Bar{A}_L^{\textit{cls}}$ obtained from $A_L^{\textit{cls}}$ on the last attention block by setting the values greater than the mean value of $A_L^{\textit{cls}}$ as 1 and the rest as 0. Cropped patches are separated into foreground patches and background patches. A patch is a foreground patch when it has more than $50\%$ of pixels belonging to $\Bar{A}_L^{\textit{cls}}$, and a background patch when it contains more than $80\%$ of pixels that belong to $(1 - \Bar{A}_L^{\textit{cls}})$. K-Means is executed on these two groups of patches separately. If a patch is treated as a foreground patch, its probability in those background categories will have a default value of 0, and a background patch is treated in a similar way.
The separation of foreground and background patches help generate more accurate semantic clusters and better pseudo labels. 

Different cropping protocols are applied to the four datasets, due to their different properties and requirements. On Cityscapes, foreground prior is not involved as there is no definition of foreground/background. On LIP, the images are obtained as bounding boxes around each person without much background area, so we treat all the areas as foreground. On COCO-80, Pascal-VOC and LIP, \eqref{eq:prob_bg} is used to generate background probability.

\PAR{Trainning Setting}. We use \textit{ViT-small} $8 \times 8$ pre-trained on ImageNet by DINO\cite{DINO} as encoder and fix the weight during training, and Segmenter~\cite{strudel2021segmenter} with random initial weights as the decoder. We pre-compute and save all the initial pseudo labels and build a pseudo label bank which considerably accelerates the training. More training details are included in the supplemental materials.

\PAR{Data Augmentation}. A set of data augmentations are utilized, such as color jittering, horizontal flipping, Gaussian blur, color dropping, and random resized crop following \cite{BYOL,DINO}. The cropped image is resized to $(\frac{H}{2}, \frac{W}{2})$ for the following training. We crop and resize the initial pre-computed pseudo label to $(\frac{h}{2}, \frac{w}{2})$ by RoI-Align operator~\cite{he2017maskrcnn}.

\subsection{Main Results}
\PAR{Baseline}. We compare our method with PiCIE~\cite{PiCIE}, IIC~\cite{IIC} and MaskContrast (MC)~\cite{MaskContrast}, which can generate segmentation masks without further fine-tuning. Since IIC and PiCIE cannot distinguish foreground objects from the background, they can only be applied on COCO-Stuff and Cityscapes. MaskContrast is only able to deal with the foreground object without background area, so it is only applied to COCO-80 and Pascal-VOC. None of these methods can actually work well on LIP which is essentially a task of fine-grained human parsing. Note that our method can be successfully adapted to all these datasets.

\begin{table}[H]
    \centering
    \small
    \caption{Results on four benchmarks. * indicates the results are evaluated on the ``curated" samples. $\dag$ denotes PiCIE trained without auxiliary clustering. \label{tab:main_results}}
    \subtable{
    \renewcommand\arraystretch{1.04}
    \begin{tabular}{ll|ll}
        \toprule
        {Dataset} & {Method} & {mIoU} & {Acc.} \tabularnewline
        \midrule
        \multirow{4}{*}{COCO-S-27*} & {IIC~\cite{IIC}} & 6.71 & 21.79\tabularnewline
        & {PiCIE$\dag$~\cite{PiCIE}} &  13.84 & 48.09 \tabularnewline
        & {PiCIE~\cite{PiCIE}} &  14.36 & 49.99 \tabularnewline
        & {TransFGU} & \bf 17.47 & \bf 52.66\tabularnewline
        \midrule
        \multirow{4}{*}{COCO-S-27} & {IIC~\cite{IIC}} & 2.36 & 21.02\tabularnewline
        & {PiCIE~\cite{PiCIE}} &  11.88 & 37.20 \tabularnewline
        & {TransFGU} & \bf 16.19 & \bf 44.52\tabularnewline
        \midrule
        \multirow{4}{*}{COCO-S-171} & {IIC~\cite{IIC}} & 0.64  & 8.67  \tabularnewline
        & {PiCIE~\cite{PiCIE}} & 4.56 & 24.66 \tabularnewline
        & {TransFGU} & \bf 11.93 & \bf 34.32\tabularnewline
        \bottomrule
    \end{tabular}}
    \subtable{
    \begin{tabular}{ll|ll}
        \toprule
        {Dataset} & {Method} & {mIoU} & {Acc.} \tabularnewline
        \midrule
        \multirow{2}{*}{COCO-80} & {MC~\cite{MaskContrast}} & 3.73 & 8.81 \tabularnewline
        & {TransFGU} & \bf 12.69 & \bf 64.31\tabularnewline
        \midrule
        \multirow{3}{*}{Cityscapes} & {IIC~\cite{IIC}} & 6.35 & 47.88 \tabularnewline
        & {PiCIE~\cite{PiCIE}} & 12.31 & 65.50 \tabularnewline
        & {TransFGU} & \bf 16.83 & \bf 77.92 \tabularnewline
        \midrule
        \multirow{2}{*}{Pascal-VOC}
        & {MC~\cite{MaskContrast}} & 35.00 & 79.84  \tabularnewline
        & {TransFGU} & \bf 37.15 & \bf 83.59  \tabularnewline
        \midrule
        LIP-5 & {TransFGU}  & 25.16 & 65.76 \tabularnewline
        LIP-16 & {TransFGU}  & 15.49 & 60.08 \tabularnewline
        LIP-19 & {TransFGU}  & 12.24 & 42.52\tabularnewline
        \bottomrule
    \end{tabular}}

\end{table}

\PAR{Quantitative evaluation.}
The comparison is shown in \tabref{tab:main_results}. Our method exceeds all the baseline methods on mIoU and pixel accuracy of all datasets. Our method has been adapted to different granularity levels in the same dataset, \eg 27/171 categories on COCO-S and 5/16/19 categories in LIP, to make a fair comparison with other methods which mainly work on coarser-level categories. Our method shows a larger margin of performance improvement for the settings with much finer-grained categories. Besides, it can also be adapted to only segment foreground objects, \ie COCO-80, Pascal-VOC, and LIP, demonstrating its all-around flexibility.

\PAR{Fair comparison with prior arts}. The difference in backbone architecture and the pre-training manner to obtain backbone weights influences the model performance. For fair comparison, we conduct the following experiments successively: (1) reproduce IIC with the ResNet-18 backbone fully-supervised trained on ImageNet to align the amount of training data; (2) reproduce IIC and PiCIE with ResNet-50 and ViT-S fully-supervised trained on ImageNet instead of its original ResNet-18 to make backbone parameters comparable with our method; (3) reproduce IIC, PiCIE and MaskContrast with the ResNet-50 backbone trained by DINO to unify the training manner; (4) reproduce the PiCIE and MaskContrast with the ViT backbone trained by DINO instead of ResNet-50 to eliminate the impact of the architecture difference. To obtain the input feature maps of FPN used in PiCIE, we extract pixel features every three attention blocks amount 12 layers and resize them to the target size. The results are shown in Table~\ref{tab:fair_comparison}. Our method outperforms all the baseline in each case. Note that the performance of IIC and PiCIE with backbones trained by SSL manner becomes inferior to its original fully-supervised version. It might be due to the feature learned by SSL containing more fine-grained foreground details, which may distract the bottom-up method to find similar features that belong to the same category.

\begin{table}[h]
    \centering
    \small
    \caption{Results for the effectiveness of pretrained weight. *, $\dag$ and $\ddag$ indicates the weights are trained on ImageNet by fully-supervised, DINO and MoCo, respectively. \label{tab:fair_comparison}}
    \begin{tabular}{cc|cc|cc|cc}
        \toprule
        {Dataset} & {Method} & {mIoU} & {Acc.} & {Dataset} & {Method} & {mIoU} & {Acc.} \tabularnewline
        \midrule
        \multirow{8}{*}{COCO-S-171} & {IIC-R18~\cite{IIC}}      & 0.64  & 8.67  & \multirow{3}{*}{COCO-S-171}   & PiCIE-R50$\dag$~\cite{PiCIE}      & 2.30      & 13.50 \tabularnewline
                                    & {IIC-R18*~\cite{IIC}}     & 1.22  & 13.92 &                               & PiCIE-ViT$\dag$~\cite{PiCIE}      & 3.02      & 18.45 \tabularnewline
                                    & IIC-R50*~\cite{IIC}       & 2.15  & 15.72 &                               & TransFGU$\dag$                    & \bf 11.93 & \bf 34.32 \tabularnewline
        \cline{5-8}                 & IIC-R50$\dag$~\cite{IIC}  & 0.98 & 11.89  & \multirow{4}{*}{Pascal-VOC}   & MC-R50$\ddag$~\cite{MaskContrast} & 35.00     & 79.84 \tabularnewline
                                    & PiCIE-R18*~\cite{PiCIE}   & 4.56 & 24.66  &                               & MC-R50$\dag$~\cite{MaskContrast}  & 28.72     & 78.72\tabularnewline
                                    & PiCIE-R50*~\cite{PiCIE}   & 5.61 & 29.79  &                               & MC-ViT$\dag$~\cite{MaskContrast}  &  31.24    & 79.18\tabularnewline
                                    & PiCIE-ViT*~\cite{PiCIE}   & 6.82 & 31.17  &                               & TransFGU$\dag$                    & \bf 37.15 & \bf 83.59 \tabularnewline
        \bottomrule
    \end{tabular}
\end{table}

\begin{figure}[h]
    \centering
    \setlength{\tabcolsep}{1pt}
    \begin{tabular}{ccccc|cccc}
        Image & IIC & PiCIE & Ours & GT & Image & MC & Ours & GT
        \tabularnewline
        \includegraphics[width=0.105\linewidth]{figures/qualitative/main/COCOStuff171/Image/000000017627.png} &
        \includegraphics[width=0.105\linewidth]{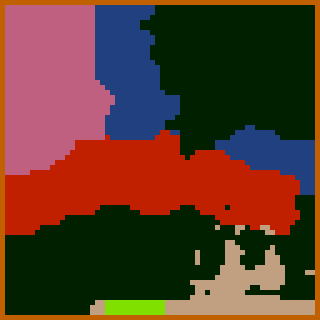} & 
        \includegraphics[width=0.105\linewidth]{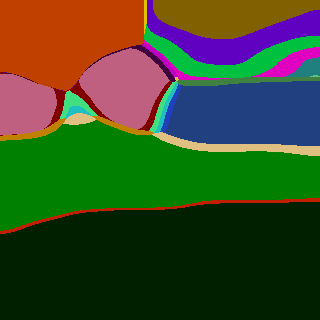} & 
        \includegraphics[width=0.105\linewidth]{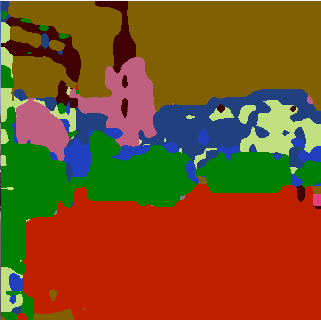} &
        \includegraphics[width=0.105\linewidth]{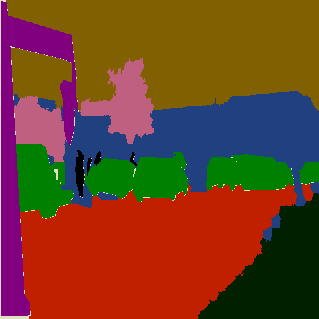} &
        \includegraphics[width=0.105\linewidth]{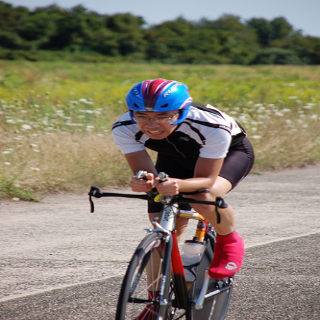} &
        \includegraphics[width=0.105\linewidth]{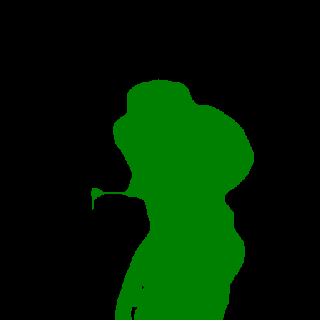} &
        \includegraphics[width=0.105\linewidth]{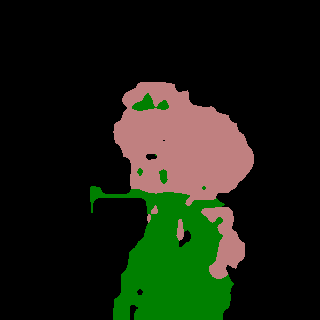} &
        \includegraphics[width=0.105\linewidth]{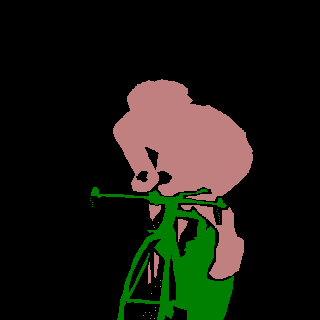}
        \tabularnewline
        \includegraphics[width=0.105\linewidth]{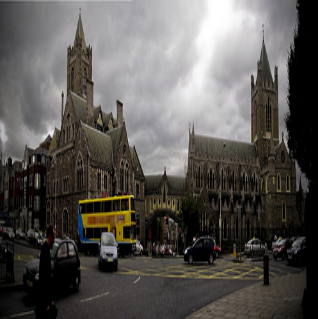} &
        \includegraphics[width=0.105\linewidth]{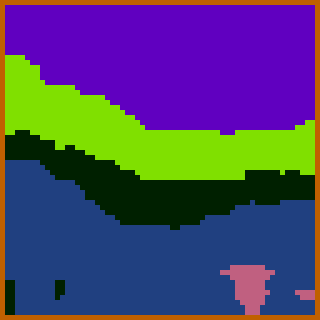} & 
        \includegraphics[width=0.105\linewidth]{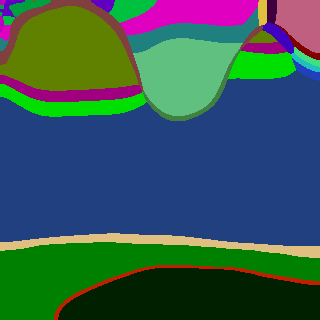} & 
        \includegraphics[width=0.105\linewidth]{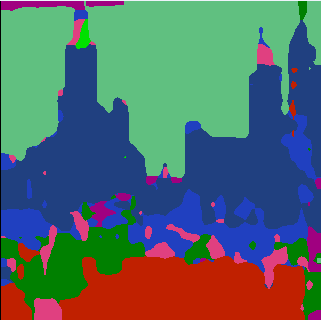} &
        \includegraphics[width=0.105\linewidth]{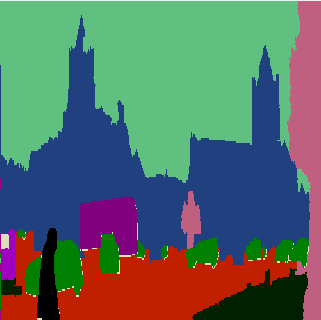} &
        \includegraphics[width=0.105\linewidth]{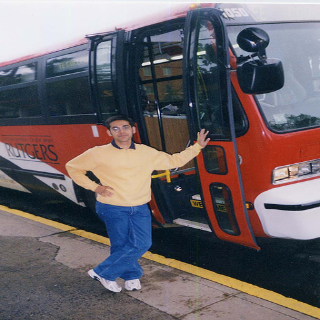} &
        \includegraphics[width=0.105\linewidth]{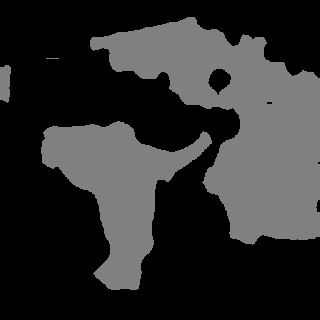} &
        \includegraphics[width=0.105\linewidth]{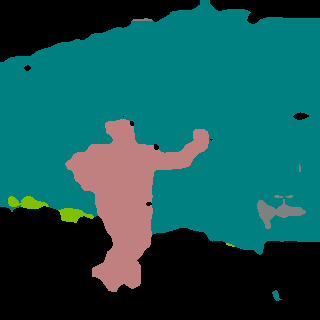} &
        \includegraphics[width=0.105\linewidth]{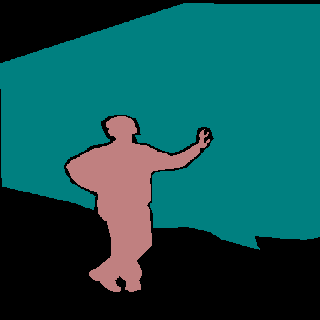}
        \tabularnewline
        \includegraphics[width=0.105\linewidth]{figures/qualitative/main/COCOStuff171/Image/000000441586.png} &
        \includegraphics[width=0.105\linewidth]{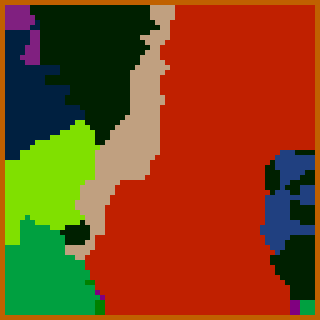} & 
        \includegraphics[width=0.105\linewidth]{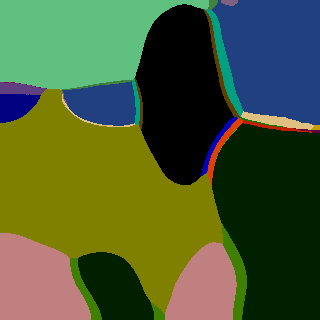} & 
        \includegraphics[width=0.105\linewidth]{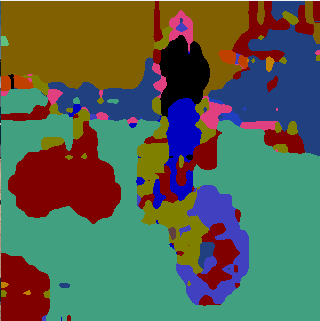} &
        \includegraphics[width=0.105\linewidth]{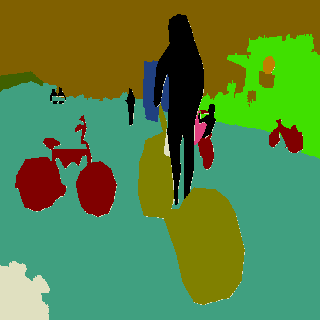} &
        \includegraphics[width=0.105\linewidth]{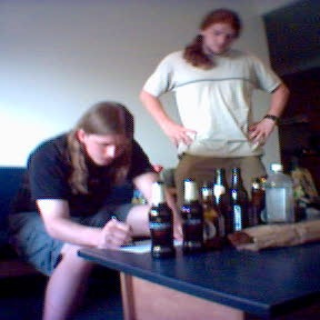} &
        \includegraphics[width=0.105\linewidth]{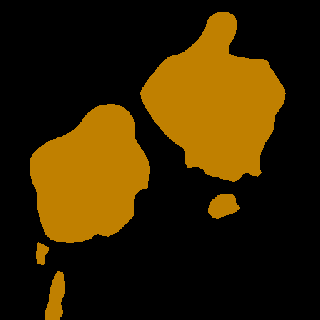} &
        \includegraphics[width=0.105\linewidth]{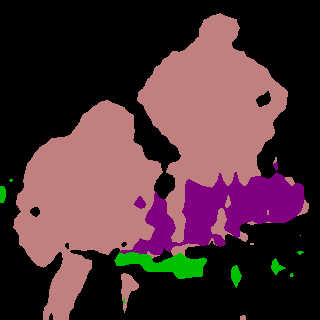} &
        \includegraphics[width=0.105\linewidth]{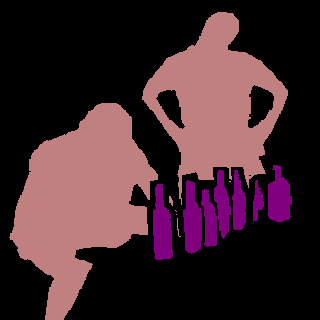}
        \tabularnewline
        \includegraphics[width=0.105\linewidth]{figures/qualitative/main/COCOStuff171/Image/000000216739.png} &
        \includegraphics[width=0.105\linewidth]{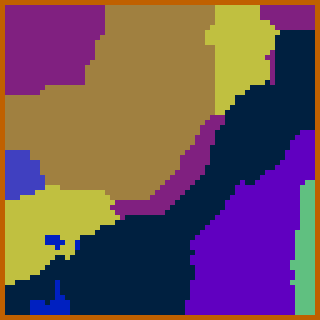} & 
        \includegraphics[width=0.105\linewidth]{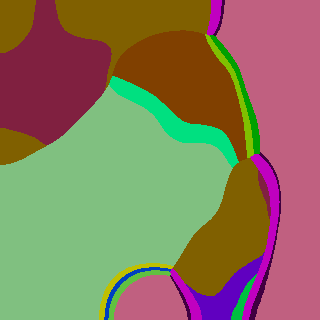} & 
        \includegraphics[width=0.105\linewidth]{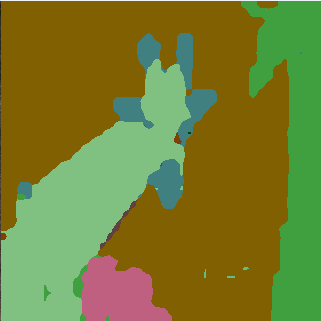} &
        \includegraphics[width=0.105\linewidth]{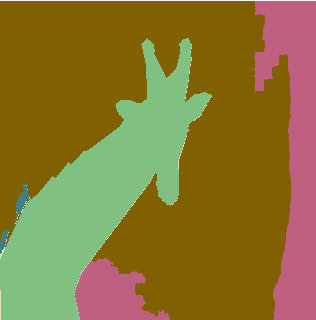} &
        \includegraphics[width=0.105\linewidth]{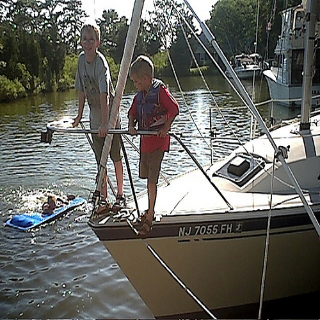} &
        \includegraphics[width=0.105\linewidth]{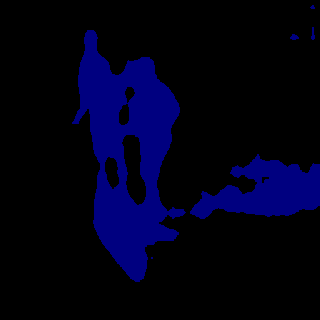} &
        \includegraphics[width=0.105\linewidth]{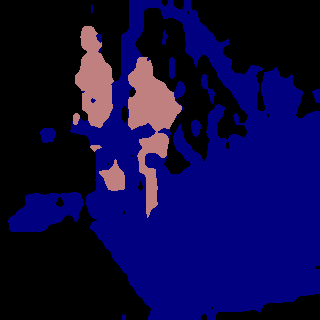} &
        \includegraphics[width=0.105\linewidth]{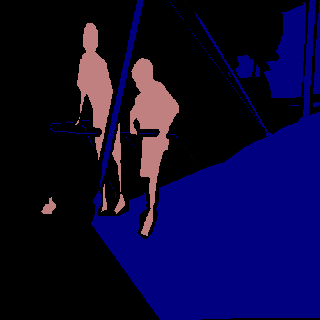}
    \end{tabular}
    \caption{Qualitative comparison on COCO-S-171 (left) and Pascal-VOC (right). \label{fig:qualitative_result}}
\end{figure}

\PAR{Qualitative evaluation.} Some qualitative comparisons on different benchmarks are shown in Figure~\ref{fig:qualitative_result}. Our results can show richer semantic information in both scene/object-centric images on all datasets. We obtain much detailed segmentation on foreground objects, especially in complicated scenarios.

\subsection{Ablation Studies}

\PAR{The Bootstrapping Mechanism}. \tabref{tab:teacher-student} compares results on MS-COCO of directly using initial pseudo labels (``initial''), training the student network once on the initial pseudo labels without the bootstrapping (``trained''), and the proposed bootstrapping mechanism (``bootstrapped''). Student network trained once on the initial pseudo labels can achieve an improved mIoU over the original pseudo labels, indicating that it can effectively learn meaningful information from the noisy pseudo label. The performance is further improved as more iterations of the teacher-student bootstrapping are introduced.

\PAR{The effectiveness of encoder and decoder}. To evaluate the effectiveness of the encoder and decoder used in our top-down pipeline, we conduct three more experiments that gradually replace the encoder and decoder with the ones used in PiCIE~\cite{PiCIE}. First, we change the encoder from ViT to ResNet-50 followed by an FPN model and keep the decoder as Segmenter. Second, we keep the encoder as ViT and change the decoder from Segmenter to a single layer convolution classifier to map the output dimension of the last attention block from $d$ to $K$. Last, the encoder and decoder are replaced by ResNet-50 and convolution classifier, respectively. In all the settings, the encoders (ViT, ResNet-50) are trained on the ImageNet by DINO, and the decoders (Segmenter, convolution classifier) are randomly initialized. The results are shown in \tabref{tab:ab:encoder_decoder}. As one can see, in all the settings, the performance can be improved from the initial pseudo label (encoder and decoder are all set to 'None'). Moreover, it's important to use fully-transformer architecture in our pipeline due to its robustness to noises.

\PAR{The Bootstrapping Loss}. We compare different combinations of using the original CE loss in~\eqref{eq:loss:ce}, peer loss in ~\eqref{eq:loss:peer}, uncertainty loss in~\eqref{eq:loss:unc}, and diversity loss in~\eqref{eq:loss:div} in \tabref{tab:ab:loss}. Each of the proposed three bootstrapping losses can effectively improve the performance of the original CE loss, and their combination achieves the best performance.

\begin{multicols}{2}
\begin{table}[H]
    \centering
    \small
    \caption{Results for the effectiveness of bootstrap mechanism on MS-COCO with various semantic levels.\label{tab:teacher-student}}
    \begin{tabular}{c|ccc}
        \toprule
        {Dataset} & {initial} & {trained} & {bootstrapped} \tabularnewline
        \midrule
        {COCO-S-27}  & 8.95 & 13.19 & 16.19  \tabularnewline
        {COCO-S-171} & 5.05 & 8.66 & 11.93  \tabularnewline
        {COCO-80} & 4.23 & 8.28 & 12.69  \tabularnewline
        \bottomrule
    \end{tabular}
\end{table}

\begin{table}[H]
    \centering 
    \small
    \caption{Results for the effectiveness of encoder and decoder on COCO-S-171.\label{tab:ab:encoder_decoder}}
    \begin{tabular}{cc|cc}
        \toprule
        {encoder} & {decoder} & {mIoU} & {Acc.} \tabularnewline
        \midrule
        None & None & 5.05 & 17.33 \tabularnewline
        R50+FPN & segmenter & 9.21 & 26.17 \tabularnewline
        ViT+FPN & classifier & 7.96 & 24.17 \tabularnewline
        R50+FPN & classifier & 8.33 & 25.35 \tabularnewline
        \bottomrule
    \end{tabular}
\end{table}

\end{multicols}

\begin{table}[H]
    \small
    \caption{Results for different losses on COCO-S-171 and Pascal-VOC.\label{tab:ab:loss}}
    \begin{tabular}{c|cccc|cc|c|cccc|cc}
        \toprule
        \multirow{2}{*}{Dataset} & \multicolumn{4}{c|}{Loss} & \multirow{2}{*}{mIoU} & \multirow{2}{*}{Acc.} & \multirow{2}{*}{Dataset} & \multicolumn{4}{c|}{Loss} & \multirow{2}{*}{mIoU} & \multirow{2}{*}{Acc.} \tabularnewline
        & CE & Peer & unc & div & & & & CE & Peer & unc & div & &  \tabularnewline
        \midrule
        \multirow{5}{*}{COCO-S-171} & \checkmark & & & & 10.49 & 29.72 & \multirow{5}{*}{Pascal-VOC} & \checkmark & & & & 34.24 & 79.85\tabularnewline
        & & \checkmark & & & 10.54 & 30.46 & & & \checkmark & & & 35.08 & 80.92\tabularnewline
        & & \checkmark & \checkmark &  & 11.24 & 33.03 & & & \checkmark & \checkmark &  & 36.46  & 82.36 \tabularnewline
        & & \checkmark & & \checkmark & 10.96 & 32.45 & & & \checkmark & & \checkmark & 35.97 & 82.03  \tabularnewline
        & & \checkmark & \checkmark & \checkmark & 11.93 & 34.32 & & & \checkmark & \checkmark & \checkmark & 37.15 & 83.59  \tabularnewline
        \bottomrule
    \end{tabular}
\end{table}

\section{Conclusion} We propose the first top-down framework for unsupervised semantic segmentation, which shows the importance of high-level semantic information in this task. The semantic prior information is learned from large-scale visual data in a self-supervised manner, and then mapped to the pixel-level feature space. By carefully designing the mapping process and the unsupervised training mechanism, we obtain fine-grained segmentation for both foreground and background. Our design also enables the flexible control of granularity levels of the semantic categories, making it possible to generate semantic segmentation results for various datasets with different requirements. The fully unsupervised manner and the flexibility make our method much more practical in real applications.

\section*{Acknowledgements} This work was supported by funds for Key R\&D Program of Hunan (2022SK2104), Leading plan for scientific and technological innovation of high-tech industries of Hunan (2022GK4010), the National Natural Science Foundation of Changsha (kq2202176), National Key R\&D Program of China (2021YFF0900602), the National Natural Science Foundation of China (61672222) and Alibaba Group through Alibaba Research Intern Program.

\clearpage
\appendix

\title{TransFGU: A Top-down Approach to Fine-Grained Unsupervised Semantic Segmentation \\ ---Supplementary Material---}

\titlerunning{TransFGU}
%
\author{Zhaoyuan Yin\inst{1,2}$^{\ast}$\thanks{
* Work done during an internship at Alibaba Group.\\
\dag~Corresponding author, project lead.\\
\ddag~Corresponding author.
}\index{Zhaoyuan Yin} 
\and Pichao Wang\inst{2\dag}
\and Fan Wang\inst{2}
\and Xianzhe Xu\inst{2}
\and \\ Hanling Zhang\inst{3\ddag}
\and Hao Li\inst{2} 
\and Rong Jin\inst{2}}
\authorrunning{Zhaoyuan Yin et al.}
%

\institute{College of Computer Science and Electronic Engineering, Hunan University, China \email{zyyin@hnu.edu.cn} \and Alibaba Group, China\\ \email{\{pichao.wang, fan.w, xianzhe.xxz, lihao.lh, jinrong.jr\}@alibaba-inc.com} \and School of Design, Hunan University, China\\ \email{jt\_hlzhang@hnu.edu.cn}\\}


\maketitle

In this supplementary material, we first present more training details. Then we present more ablation studies to reveal the effectiveness of K, the semantic prior model used in our pipeline, the cropping manner and the foreground prior. Our re-defined labels for 5 and 16 categories in LIP are also presented. Finally, we show failure cases, the visualization of CAM and more qualitative results on Cityscapes and LIP.

\PAR{Training Details}. We train our model on 4 $\times$ Tesla V100 32G GPUs with Adam optimizer. In general, a larger batch-size benefits the learning stability, and a larger amount of images and target classes requires a larger learning rate and training epochs. Different learning rates, batch-size, and epochs are used for the four benchmarks for a better trade-off between the performance and the GPU memory cost. Specifically, we set 256/1e-4/200 (batch size/learning rate/epochs) for MS-COCO, 512/1e-4/100 for PascalVOC, 256/5e-5/50 for Cityscapes and 256/2e-5/100 for Lip. The learning rate is fixed during the whole training process. The $\alpha$ in peer loss linearly increases from $0.03$ to $0.1$ according to the number of training epochs. The background threshold $T^{bg}$, loss weights $\omega 1, \omega 2$, and the parameter $\beta$ in the step size of sliding window is fixed across all the benchmarks.

\PAR{The Effectiveness of K}. We perform overclustering on the Pascal-VOC to reveal the effectiveness of different settings of K to a certain target semantic granularity level. 
Specifically, we set different $K$ which larger than the number of ground-truth categories as the target amount of cluster when performing K-Means to the top-level semantic features, and use the Hungarian algorithm to find the greatest matching. Note that the unmatched categories are discarded when calculating the mIoU and accuracy. The results shown in \tabref{tab:overclustering} indicate a reasonable $K$ value that closed to the amount of ground-truth categories can benefit the evaluation performance, while an over-large $K$ leads to the performance drop. It is due to the over-clustering in top-level semantic features leads to a finer semantic division that would divide a whole object into its sub-categories, and causes the mismatching between predicted and the ground-truth semantic clusters. We emphasize that trying to find an 'objective' value of K concerning a certain validation set in an unsupervised way is difficult. It’s due to the intrinsic non-uniqueness and hierarchy of semantic definition, e.g. both ‘left/right arm’ or ‘arms’ are acceptable semantic divisions on the same validation set, depending on which granularity level is required. To this end, the introduction of K in our pipeline is a necessary and reasonable way to enable flexible control on any desired granularity levels.

\begin{multicols}{2}

\begin{table}[H]
    \centering
    \small
    \caption{Over-clustering Results on Pascal-VOC (20 ground-truth categories) with different setting of $K$.\label{tab:overclustering}}
    \begin{tabular}{c|ccccc}
        \toprule
        {K} & {20} & {22} & {25} & {30} & {50} \tabularnewline
        \midrule
        {mIoU}  & 37.15 & 37.26 & 37.68 & 29.39 & 27.42  \tabularnewline
        {Acc}  & 83.59 & 84.12 & 82.65 & 80.85 & 75.32 \tabularnewline
        \bottomrule
    \end{tabular}
\end{table}

\begin{table}[H]
    \centering 
    \small
    \caption{Results for the effectiveness of semantic prior model.\label{tab:ab:semantic_prior}}
    \begin{tabular}{cc|cc}
        \toprule
        {SSL method} & {patch size} & {mIoU} & {Acc.} \tabularnewline
        \midrule
        DINO & 8 & 11.93 & 34.32 \tabularnewline
        DINO & 16 & 9.43 & 28.12 \tabularnewline
        MoCoV3 & 16 & 2.32 & 20.94 \tabularnewline
        \bottomrule
    \end{tabular}
\end{table}


\end{multicols}

\PAR{The Effectiveness of the Semantic Prior Model}. We conduct two more experiments to reveal the effectiveness of the semantic prior model. The results are shown in \tabref{tab:ab:semantic_prior}. We first change the semantic prior model from ViT small 8x8 to ViT small 16x16 trained by DINO~\cite{DINO}. The mIoU drop from $11.93$ to $9.43$. Second, we change the ViT small 16x16 trained by DINO with the one trained by MoCo V3~\cite{MoCoV3}. The performance drops dramatically to $2.32$, which indicates the different manners of self-supervised learning affect the segmentation property of the semantic prior model and result in different performances in our pipeline.

\PAR{The Quality of Cropping}. An extra experiment is conducted by using ground-truth labels instead of sliding windows to crop the foreground and background objects in a more precise way, which can be regarded as an upper limit of our cropping operation. The mIoU on COCO-Stuff-171 is slightly improved from $11.93$ to $12.16$. This shows that, even though the sliding window cropping cannot generate precise boxes around semantic objects, it still has exploited sufficient semantic information, leaving not much room to improve compared with the ground-truth cropping.

\PAR{The Foreground Prior}. Experiments of clustering the semantic features without separating the patches with foreground prior will result in the drop of mIoU from $11.93$ to $9.52$ on COCO-Stuff-171, indicating that the separation of foreground and background patches is helpful to improve the quality of semantic feature clustering.


\PAR{Label Definition on LIP under Different Granularity}. We re-defined the LIP~\cite{gong2017lip} under different granularity by merging categories in its original label (19 categories) and generating two new label definitions (16 and 5 categories). The indices projection is shown in \tabref{tab:LIP_difinitions}.

\begin{table}[H]
    \centering 
    \small
    \caption{Label definition on LIP under different granularity.\label{tab:LIP_difinitions}}
    \begin{tabular}{c|ccc}
        \toprule
        \multirow{2}{*}{Name} & \multicolumn{3}{c}{granularity level}\tabularnewline
        & {19 categories} & {16 categories} & {5 categories} \tabularnewline
        \midrule
        Background & 0 & 0 & 0 \tabularnewline
        Hat & 1 & 1 & 1 \tabularnewline
        Hair & 2 & 2 & 1 \tabularnewline
        Glove & 3 & 3 & 3 \tabularnewline
        Sunglasses & 4 & 4 & 1 \tabularnewline
        Upper-clothes & 5 & 5 & 2 \tabularnewline
        Dress & 6 & 6 & 2 \tabularnewline
        Coat & 7 & 7 & 2 \tabularnewline
        Socks & 8 & 8 & 5 \tabularnewline
        Pants & 9 & 9 & 4 \tabularnewline
        Jumpsuits & 10 & 10 & 2 \tabularnewline
        Scarf & 11 & 11 & 2 \tabularnewline
        Skirt & 12 & 12 & 2 \tabularnewline
        Face & 13 & 13 & 1 \tabularnewline
        Left-arm & 14 & 14 & 3 \tabularnewline
        Right-arm & 15 & 14 & 3 \tabularnewline
        Left-leg & 16 & 15 & 4 \tabularnewline
        Right-leg & 17 & 15 & 4 \tabularnewline
        Left-shoe & 18 & 16 & 5 \tabularnewline
        Right-shoe & 19 & 16 & 5 \tabularnewline
        \bottomrule
    \end{tabular}
\end{table}

\begin{figure}[H]
    \small
    \centering
    \setlength{\tabcolsep}{1pt}
    \begin{tabular}{cccccc}
        Image & Prediction & GT & Image & Prediction & GT
        \tabularnewline
        \includegraphics[width=0.135\linewidth]{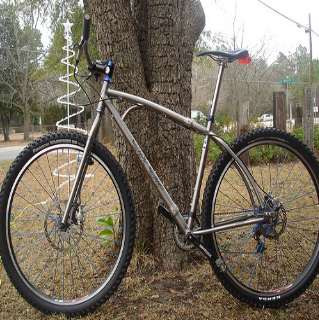} &
        \includegraphics[width=0.135\linewidth]{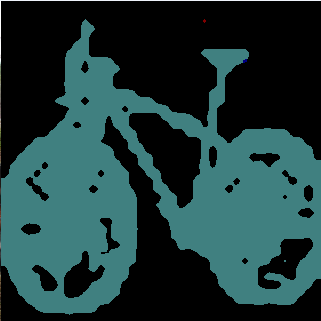} &
        \includegraphics[width=0.135\linewidth]{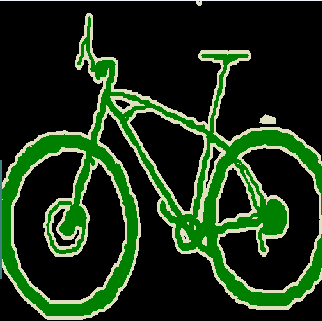} &
        \includegraphics[width=0.135\linewidth]{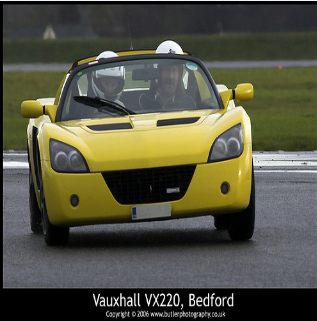} &
        \includegraphics[width=0.135\linewidth]{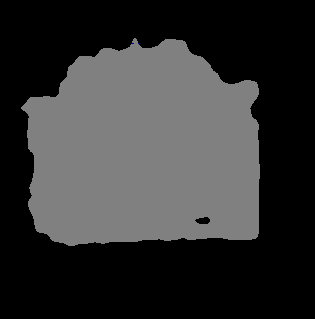} &
        \includegraphics[width=0.135\linewidth]{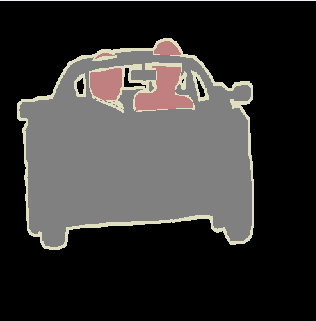}
        \tabularnewline
        \includegraphics[width=0.135\linewidth]{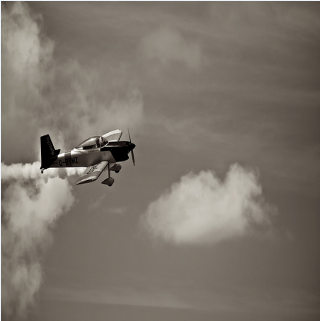} &
        \includegraphics[width=0.135\linewidth]{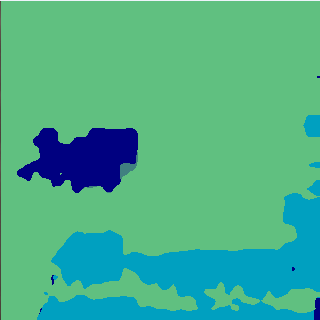} &
        \includegraphics[width=0.135\linewidth]{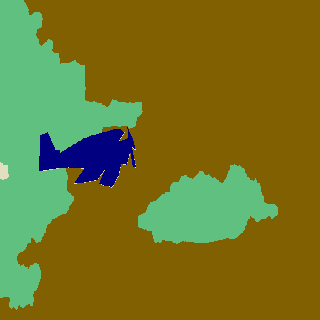} &
        \includegraphics[width=0.135\linewidth]{figures/qualitative/FailureCase/Image/000000300659.png} &
        \includegraphics[width=0.135\linewidth]{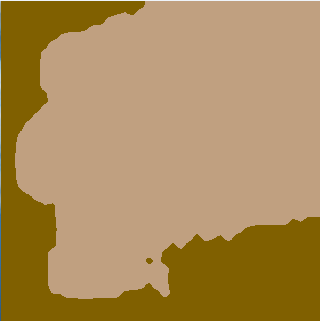} &
        \includegraphics[width=0.135\linewidth]{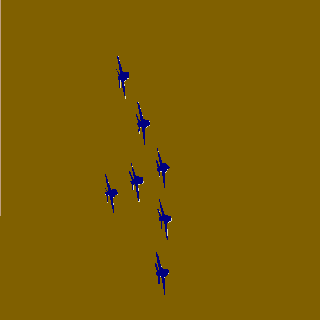}
    \end{tabular}
    \caption{Failure Cases. left: misclassified; right: missegmented. \label{fig:failure_case}}
\end{figure}

\PAR{Failure Cases}. Since our segmentation is based on the high-level semantic representation,  similar categories may confuse the classification (The two cases in the left column of \figref{fig:failure_case}). Besides that, other challenging cases would also cause failure of our method, \eg occlusions and tiny objects (top-right and bottom-right in \figref{fig:failure_case}).

\begin{figure}[H]
    \small
    \centering
    \setlength{\tabcolsep}{1pt}
    \begin{tabular}{cccc}
        Image & Bicycle & Person & Car
        \tabularnewline
        \includegraphics[width=0.200\linewidth]{figures/qualitative/CAM/Image/001.png} &
        \includegraphics[width=0.200\linewidth]{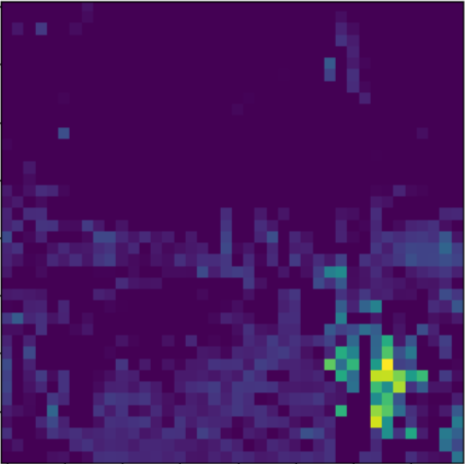} &
        \includegraphics[width=0.200\linewidth]{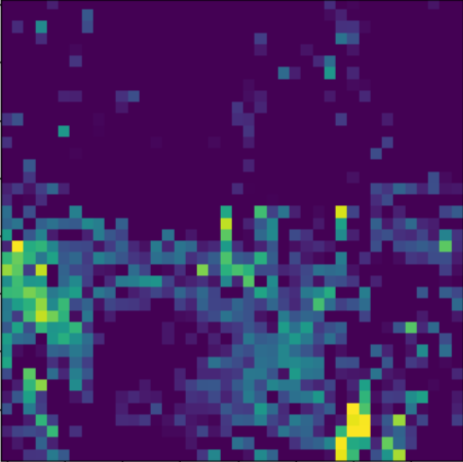} &
        \includegraphics[width=0.200\linewidth]{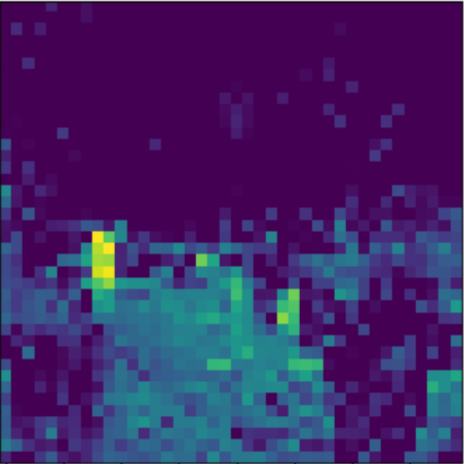}
        \tabularnewline
        \includegraphics[width=0.200\linewidth]{figures/qualitative/CAM/Image/002.png} &
        \includegraphics[width=0.200\linewidth]{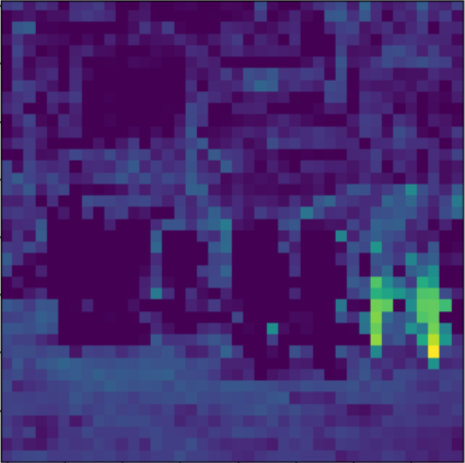} &
        \includegraphics[width=0.200\linewidth]{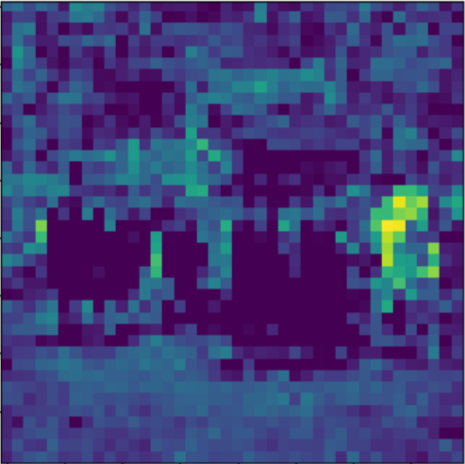} &
        \includegraphics[width=0.200\linewidth]{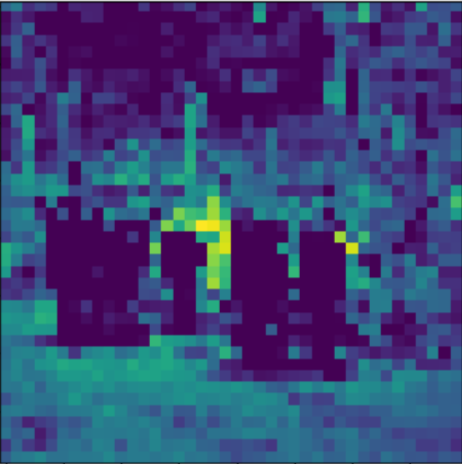}
    \end{tabular}
    \caption{CAM visualization. We show three CAMs correspondence to three particular semantic classes, \ie bicycle, person and car in the same image.\label{fig:grad_cam}}
\end{figure}

\PAR{CAM Visualization}. \figref{fig:grad_cam} visualizes the class activate map generated by Grad-CAM.  As one can see, the CAM can locate objects well in a very complex scene-based image and give a fine-grained activate map of the target location, which is crucial to our top-down segmentation pipeline.

\begin{figure}[H]
    \small
    \centering
    \setlength{\tabcolsep}{1pt}
    \begin{tabular}{ccccc}
        Image & IIC & PiCIE & TransFGU & GT
        \tabularnewline
        \includegraphics[width=0.190\linewidth]{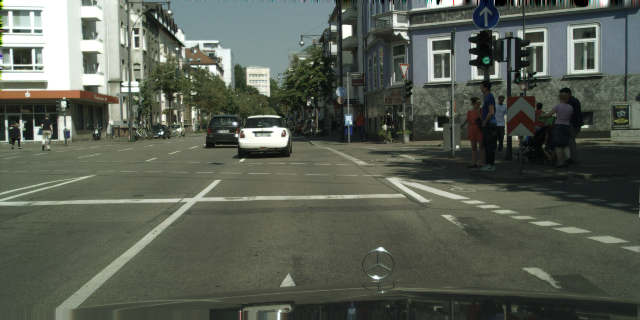} &
        \includegraphics[width=0.190\linewidth]{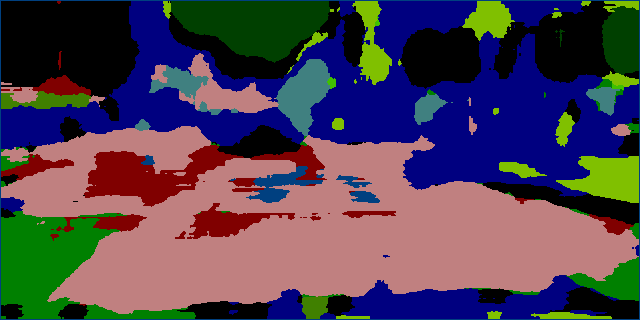} &
        \includegraphics[width=0.190\linewidth]{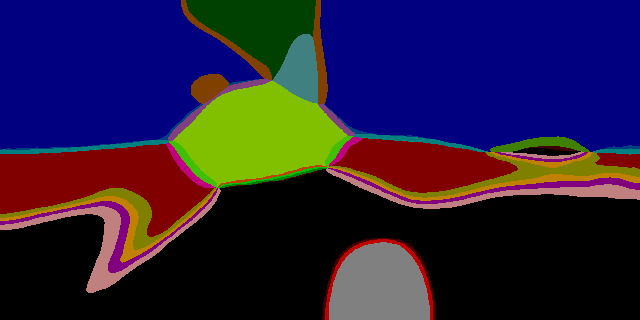} &
        \includegraphics[width=0.190\linewidth]{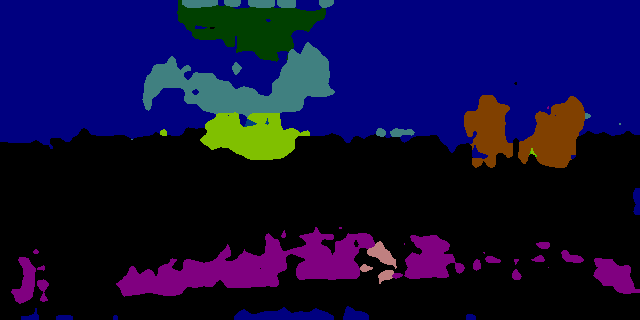} &
        \includegraphics[width=0.190\linewidth]{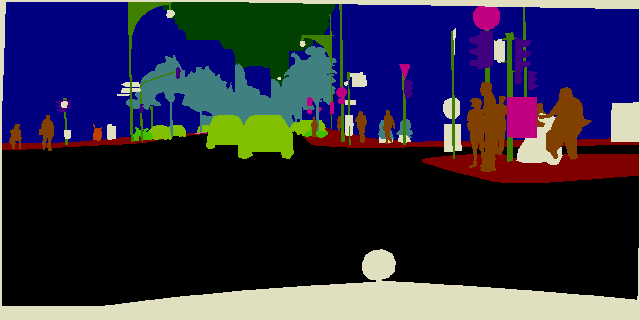}
        \tabularnewline
        \includegraphics[width=0.190\linewidth]{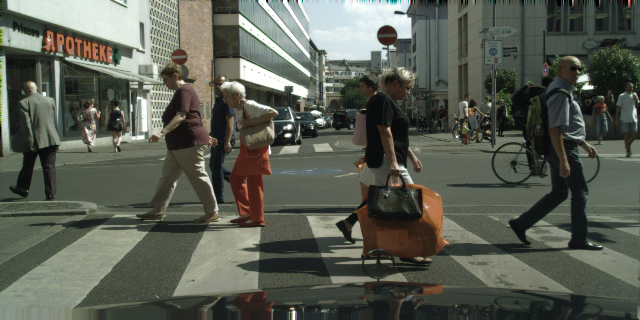} &
        \includegraphics[width=0.190\linewidth]{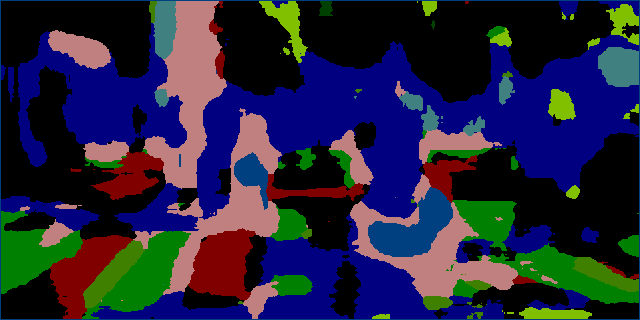} &
        \includegraphics[width=0.190\linewidth]{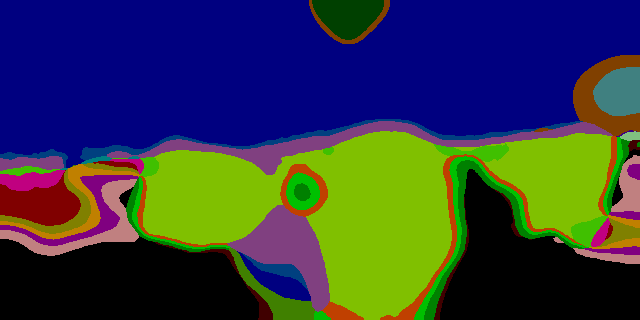} &
        \includegraphics[width=0.190\linewidth]{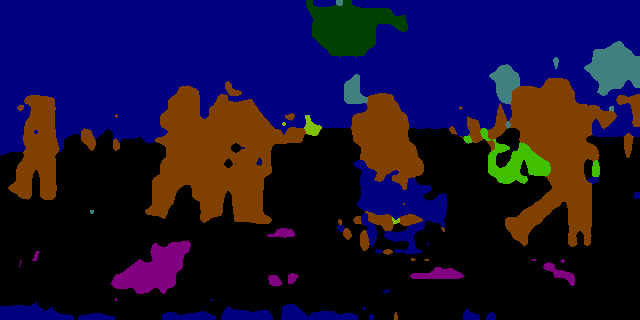} &
        \includegraphics[width=0.190\linewidth]{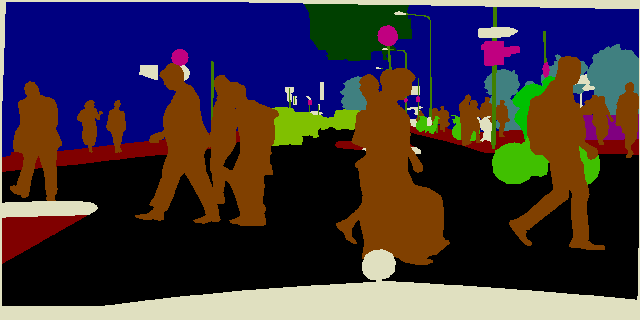}
        \tabularnewline
        \includegraphics[width=0.190\linewidth]{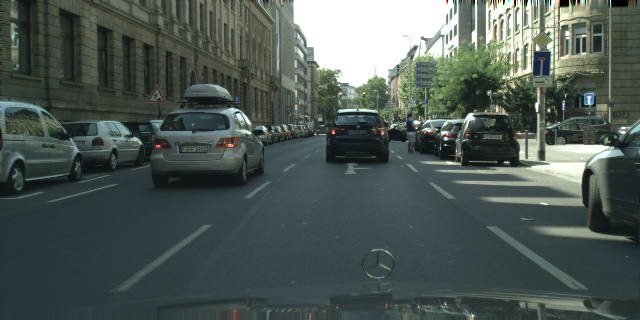} &
        \includegraphics[width=0.190\linewidth]{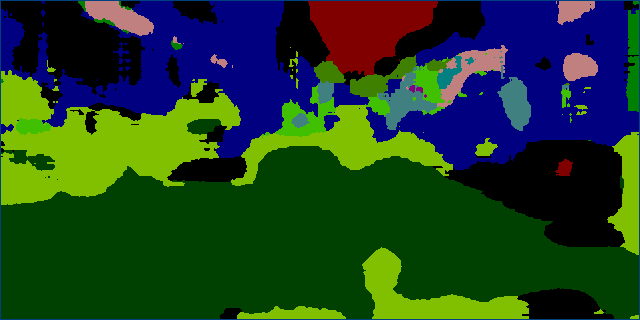} &
        \includegraphics[width=0.190\linewidth]{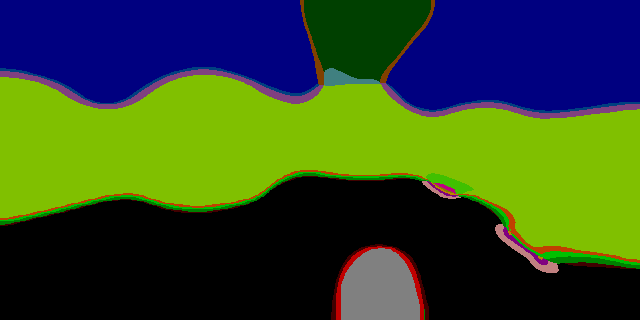} &
        \includegraphics[width=0.190\linewidth]{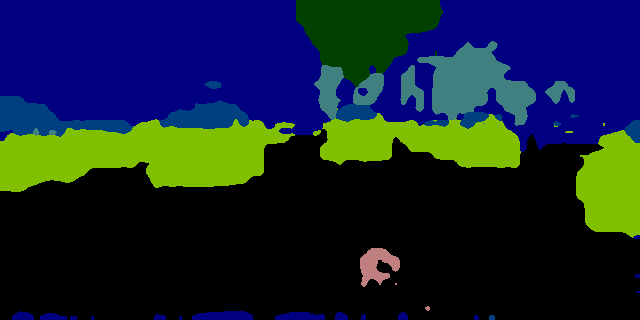} &
        \includegraphics[width=0.190\linewidth]{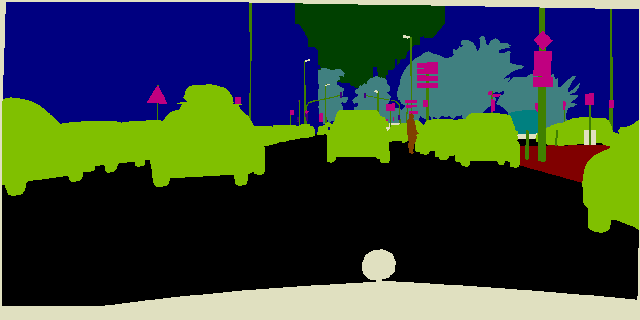}
        \tabularnewline
        \includegraphics[width=0.190\linewidth]{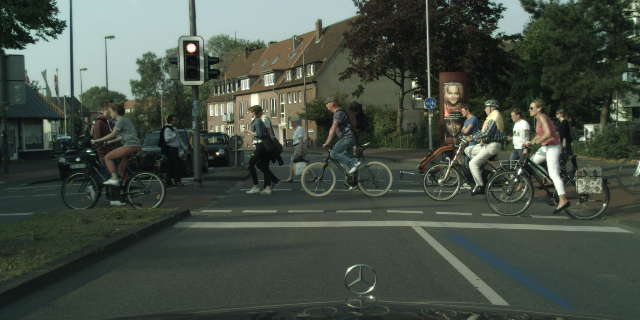} &
        \includegraphics[width=0.190\linewidth]{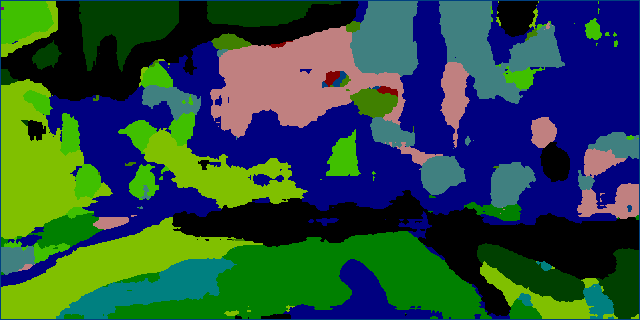} &
        \includegraphics[width=0.190\linewidth]{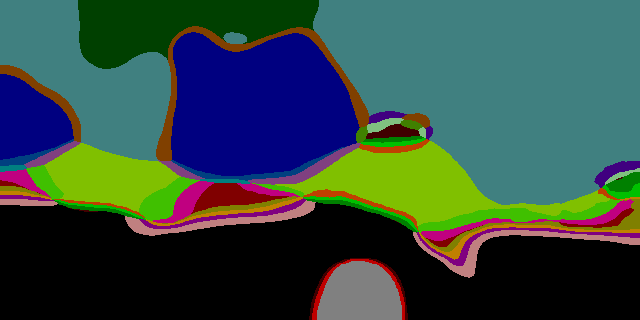} &
        \includegraphics[width=0.190\linewidth]{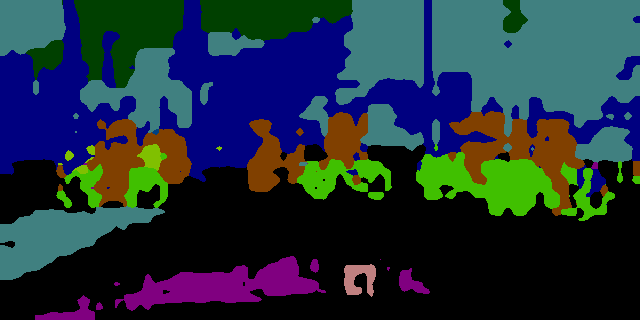} &
        \includegraphics[width=0.190\linewidth]{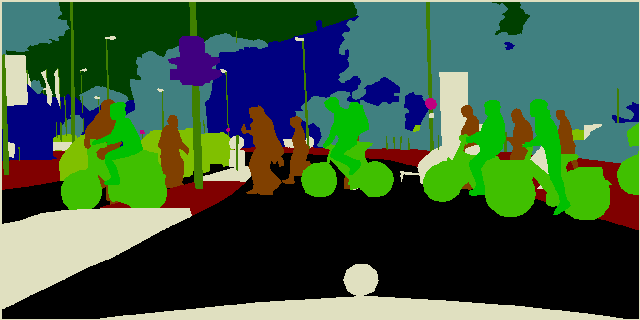}
    \end{tabular}
    \caption{Qualitative comparison on Cityscapes. \label{fig:cityscapes_result}}
\end{figure}

\begin{figure}[H]
    \small
    \centering
    \setlength{\tabcolsep}{1pt}
    \begin{tabular}{cccccc}
        Image & Pred & GT & Image & Pred & GT
        \tabularnewline
        \includegraphics[width=0.135\linewidth]{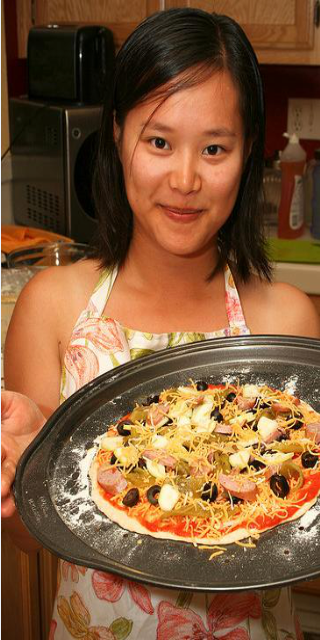} &
        \includegraphics[width=0.135\linewidth]{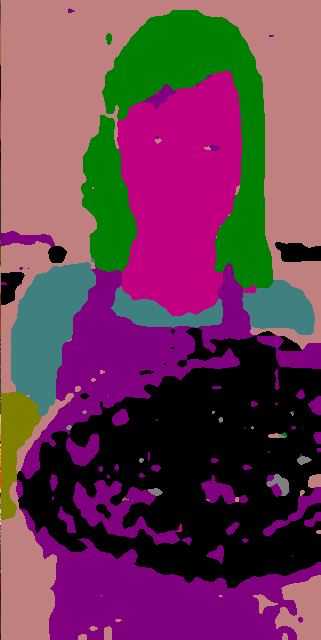} &
        \includegraphics[width=0.135\linewidth]{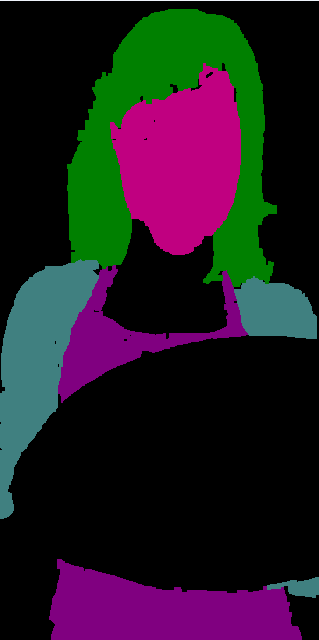} &
        \includegraphics[width=0.135\linewidth]{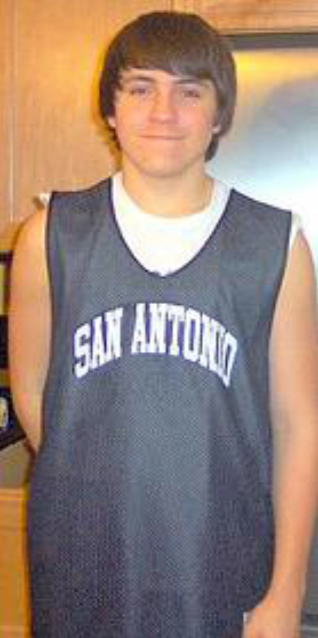} &
        \includegraphics[width=0.135\linewidth]{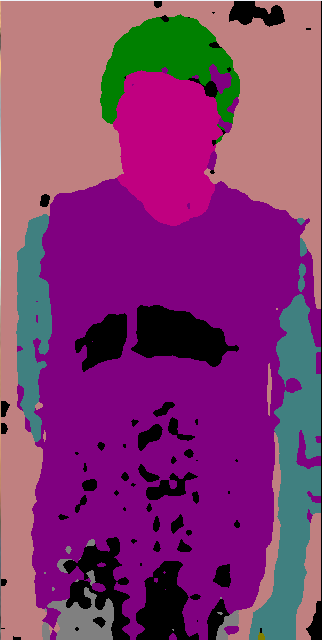} &
        \includegraphics[width=0.135\linewidth]{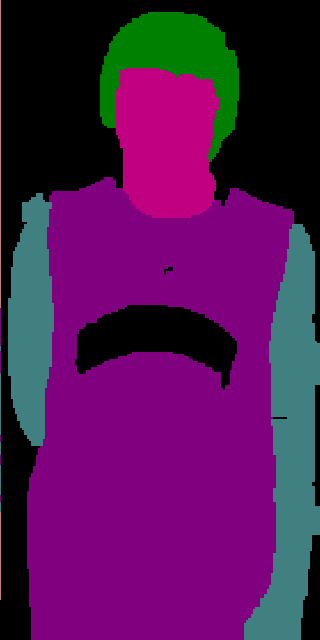}
        \tabularnewline
        \includegraphics[width=0.135\linewidth]{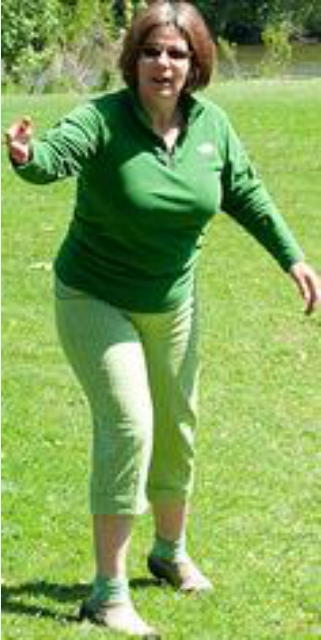} &
        \includegraphics[width=0.135\linewidth]{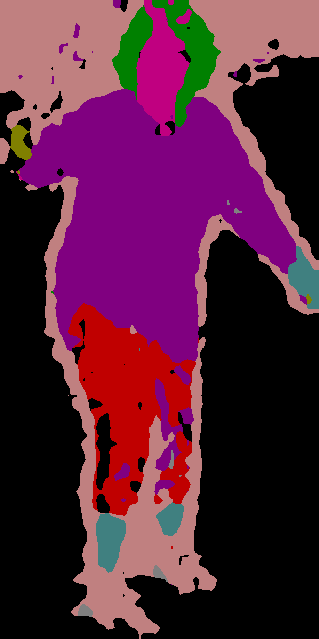} &
        \includegraphics[width=0.135\linewidth]{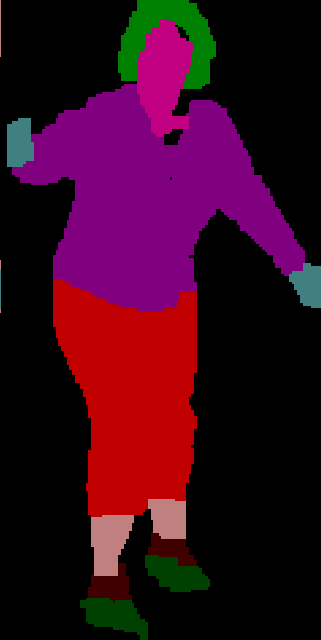} &
        \includegraphics[width=0.135\linewidth]{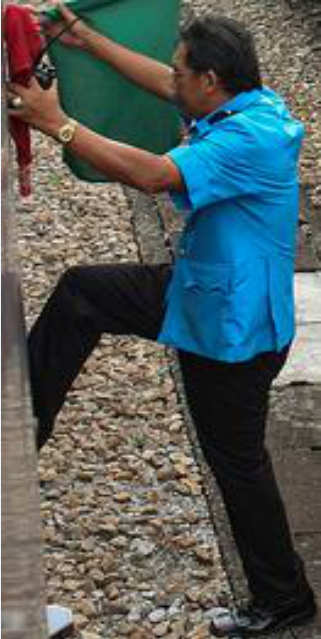} &
        \includegraphics[width=0.135\linewidth]{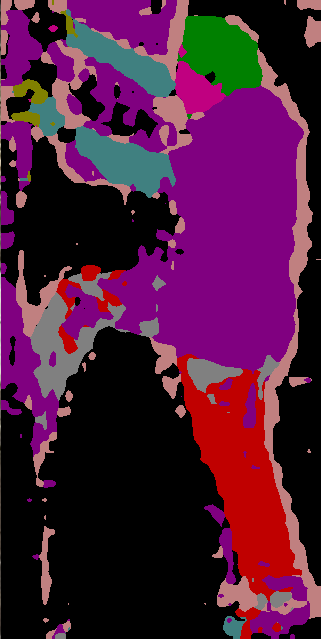} &
        \includegraphics[width=0.135\linewidth]{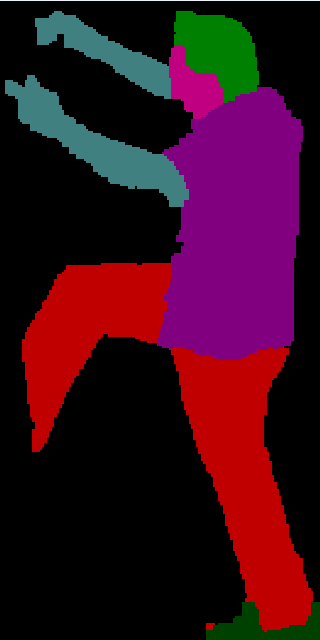}
    \end{tabular}
    \caption{Qualitative results on LIP for 16 fine-grained semantic granularity.\label{fig:LIP_result}}
\end{figure}

\PAR{Qualitative Results}. We show more qualitative results on Cityscapes~\cite{cordts2016cityscapes} and LIP~\cite{gong2017lip} in ~\figref{fig:cityscapes_result} and ~\figref{fig:LIP_result}, respectively.

%
%
\bibliographystyle{splncs04}
\bibliography{eccv}

\end{document}